\pdfoutput=1

\documentclass[journal]{IEEEtran}

\usepackage{amssymb}
\usepackage{amstext}
\usepackage{amsmath}
\usepackage{amsthm}
\usepackage{epsfig}
\usepackage{epstopdf}
\usepackage{epstopdf}
\usepackage{calc}
\usepackage{color}
\usepackage{graphicx}
\usepackage{caption}
\usepackage{booktabs}
\usepackage{rotating}
\usepackage{pdflscape}
\usepackage{slashbox}
\usepackage{subfig}

\begin{document}

\title{Modeling The Stable Operating Envelope For Partially Stable Combustion Engines Using Class Imbalance Learning}

\author{Vijay Manikandan Janakiraman,
        XuanLong Nguyen,
        Jeff Sterniak,
        and Dennis Assanis 
\thanks{Vijay Manikandan Janakiraman is with the Department of Mechanical Engineering, University of Michigan, Ann Arbor, MI, USA e-mail: vijai@umich.edu.}
\thanks{XuanLong Nguyen is with the Department of Statistics, University of Michigan, Ann Arbor, MI, USA.}
\thanks{Jeff Sterniak is with Robert Bosch LLC, Farmington Hills, MI, USA.}
\thanks{Dennis Assanis is with the Stony Brook University, NY, USA.}}

\markboth{IEEE TRANSACTIONS, In review}%
{Shell \MakeLowercase{\textit{et al.}}: Modeling The Stable Operating Envelope For Partially Stable Combustion Engines Using Class Imbalance Learning}

\maketitle

\begin{abstract}
Advanced combustion technologies such as homogeneous charge compression ignition (HCCI) engines have a narrow stable operating region defined by complex control strategies such as exhaust gas recirculation (EGR) and variable valve timing among others. For such systems, it is important to identify the operating envelope or the boundary of stable operation for diagnostics and control purposes. Obtaining a good model of the operating envelope using physics becomes intractable owing to engine transient effects. In this paper, a machine learning based approach is employed to identify the stable operating boundary of HCCI combustion directly from experimental data. Owing to imbalance in class proportions in the data, two approaches are considered. A re-sampling (under-sampling, over-sampling) based approach is used to develop models using existing algorithms while a cost-sensitive approach is used to modify the learning algorithm  without modifying the data set. Support vector machines and recently developed extreme learning machines are used for model development and results compared against linear classification methods show that cost-sensitive versions of ELM and SVM algorithms are well suited to model the HCCI operating envelope. The prediction results indicate that the models have the potential to be used for predicting HCCI instability based on sensor measurement history.
\end{abstract}



\section{Introduction} \label{intro}
HCCI engines have been studied in the last decade owing to their ability to reduce emissions and fuel consumption significantly compared to traditional spark ignition and compression ignition engines \cite{thring1,Christensen2,Aoyama3}. The highly efficient operation of HCCI is achieved using advanced control strategies such as exhaust gas recirculation (EGR) \cite{kyoungjoon}, variable valve timings (VVT) \cite{Johansson2010}, intake charge heating \cite{heatedintake} among others. However, complex manipulation of the system results in a highly nonlinear behavior \cite{nonlin_HCCI} and narrow region of stable operation \cite{narrow1,narrow2}.

In order to develop controllers and operate the system in a stable manner, it is imperative that the operating envelope of the system be determined. In general, the operating envelope can be defined as a region in the input space (of permissible values of system actuators for different thermal conditions of the engine) that results in a stable operation of the engine. Knowledge of the operating envelope is crucial for designing efficient controllers for the following reasons. The developer can get insights on the actuator extremes (for example, minimum and maximum fuel injection rates at a given speed and load condition) of the engine especially during transients \cite{kyoungjoon}. Such information can be used to enforce constraints on the control variables for desired engine operation. Also, the operating envelope model can act as a filter to perform system identification by eliminating excitations that might lead the system to be unstable. Further, the model can be used to alarm the onboard diagnostics if the engine is about to misfire \cite{misfire_emissions} owing to changes in system or operating conditions.

HCCI engines are very complex systems involving chemical kinetics and thermal dynamics which requires high-fidelity modeling using numerical simulations for capturing accurate combustion behavior \cite{Kong20071483,Zheng2009354,Wang20061831,Yao2009398}. Such an approach is computationally expensive particularly when there are several influencing variables. The situation is worsened by the transient effects, i.e., the variables along with its time history affects the system behavior, a characteristic of dynamic systems. The operating envelope that depends on the system variables along with its time history \cite{kyoungjoon} can be considered a dynamic system, and capturing this time varying behavior using conventional methods becomes intractable. Hence an approach using machine learning is considered in this paper where time series data from the sensors can be used to model the operating envelope of the HCCI engine.

The problem of identifying the operating envelope using experimental data reduces to a binary classification problem. Experimental data from the engine sensors can be labeled as belonging to stable or unstable operation and a decision boundary can be modeled using existing classification algorithms. A support vector machine algorithm was used to identify the operating envelope of a GDI engine \cite{svm_ilya} but the boundary was assumed to be a static system and time history was not considered resulting in a simple binary classification problem. However, for HCCI engines whose combustion behavior is influenced by EGR from previous cycles, the importance of considering the time history of measurements becomes important \cite{kyoungjoon}. Designing a classifier based on dynamic HCCI data is one of the objectives of this paper. Also, the experimental data consists of a large set of stable class data with limited unstable class data, as misfiring the engine is undesirable for the emission control hardware, a regular classification might not be an appropriate solution since the decision boundary would be biased to one class of data, resulting in over-fitting. Therefore, the other objective of this paper is to perform imbalanced class learning on the HCCI engine data. The following two approaches have been compared
\begin{enumerate}
  \item Heuristic re-sampling of data: apply preprocessing methods such as under-sampling and over-sampling of data to get a balanced data set.
  \item Cost-sensitive approach: modify the objective function of the learning system to weigh the minority class data more heavily.
\end{enumerate}
Three classification models including support vector machines (SVM), extreme learning machines (ELM) and logistic regression (LR) have been applied for the HCCI boundary learning problem. The models are compared for generalization accuracy, memory requirement in terms of number of parameters used to store the model and potential for online learning.

The paper is organized as follows. A brief background on the classification algorithms along with cost-sensitive modifications is given in section \ref{preliminaries}. The HCCI engine experiments and data processing are briefed in section \ref{system} with the envelope modeling and prediction results discussed in section \ref{modeling} followed by conclusions in section \ref{concl}.

\section{Classification Algorithms} \label{preliminaries}
Consider the data set $\{(x_1,y_1),...,(x_N,y_N)\}\in \big(\mathcal{X},\mathcal{Y}\big)$ where $\mathcal{X}$ denotes the space of the input features (let $\mathcal{X} = \mathbb{R}^n$) while $\mathcal{Y}$ takes values in \{-1,+1\} and $N$ denotes the number of observations. The goal of the classification algorithm is to model the underlying boundary separating the data by minimizing a risk function $R(w)$ with respect to the model parameters $w$.
\begin{equation}\label{risk}
R(w)=\frac{1}{N}\sum_{i=1}^{N}L(y_i-\hat{y_i}(x|w))+\frac{1}{2}w^Tw
\end{equation}
Here, $R(w)$ has two components - the empirical risk minimizing the training error and the structural risk minimizing the model parameters, $L$ represents a loss function and $\hat{y}(x|w,b)$ represents the model prediction, whose structures are given by the learning algorithm (see following subsections). The algorithms considered in this study are logistic regression, support vector machines and extreme learning machines. Each of the algorithm is unique in its formulation, loss function used, convergence rates, computation demand, prediction accuracy and potential for online learning. However, the main criteria used for evaluation in this study are prediction accuracy, number of parameters used for modeling and potential for online implementation for the HCCI engine system. The HCCI classification problem involves identifying the boundary separating the input space that result in a stable or unstable operation. Also, when the engine misfires, the excitation command is changed to attempt a stable operation \cite{vijay_springer}. This results in a class imbalance learning problem as the number of unstable class data is significantly smaller than the number of stable class data.

\subsection{Class Imbalance Learning}
Class imbalance learning (CIL) is encountered during situations when the number of instances of one class is very different from the number of instances in others. In a binary classification problem, the class where the number of observations are large in number is referred to as the majority class (labeled +1) while the other class is referred to as the minority class (labeled -1). Imbalanced data sets need careful attention, as machine learning (typically an optimization problem) causes the decision boundary to be favorable to the majority class data while ignoring the minority class data \cite{imbalanced_learning,svm_imbalanced}.

Several solutions have been proposed to handle CIL problems including re-sampling the data where the minority class can be duplicated to be in proportion with the majority class (referred to as over-sampling) or some majority class data is removed to match proportions with the minority class (referred to as under-sampling). Although both sampling methods aim to artificially obtain a balanced data set, under-sampling is prone to loss of majority class information while over-sampling is prone to over-fitting \cite{imbalanced_learning,svm_imbalanced}. Algorithm level modifications are also common which include cost-sensitive learning that weights the minority class data more than the majority class data in the optimization objective function. Other methods such as adjusting the decision threshold, one-class learning etc. are available in literature, but the focus of this paper is the comparison of under-sampling, over-sampling and cost-sensitive methods.

\subsection{Logistic Regression}
Logistic regression (LR) is a classical linear classifier that proves to be effective especially for large data set problems owing to its computational efficiency. LR makes use of a logistic function given by equation \eqref{logistic} which confines the output of the function to lie between 0 and 1. Unlike the linear regression model which solves a least squares problem with a squared loss function, LR solves a nonlinear optimization problem using a logistic loss function (see Figure \ref{app:loss_fig} in appendix \ref{app:loss}). The logistic loss function is particularly attractive for classification because the algorithm does not penalize the correctly classified points (at large positive margin in Figure \ref{app:loss_fig}) as much as the squared loss improving convergence.
\begin{equation}\label{logistic}
\psi(x)=\frac{1}{1+e^{-x}}
\end{equation}
The conditional probability of estimating $y$ from $x$ can be expressed in terms of the model parameters $\beta=[\beta_0 \quad \beta_1]^T$ as
\begin{equation}\label{LR_model}
P(Y=y|X=x)=\frac{1}{1+e^{-y(\beta_1^T x+\beta_0)}}
\end{equation}
where $X$ and $Y$ represent the input and output random variables. The goal of logistic regression is to determine $\beta$ such that $P(Y|X,\beta)$ is maximized using the following optimization problem (see appendix \ref{app:logreg})
\begin{equation}\label{likelihood}
\beta^* = \arg\min_\beta \sum_{i=1}^N \log  \left(1+e^{-y(\beta_1^T x+\beta_0)}\right)
\end{equation}
The equation \eqref{likelihood} is nonlinear in $\beta$ and can be solved by simple iterative methods \cite{dobson_book}. The LR decision hypothesis is given by
\begin{equation}\label{LR_hyp}
f(x)=sgn(\beta_1^T x+\beta_0)
\end{equation} where
\begin{equation}\label{}
 sgn(x) =
  \begin{cases}
   1 &  x > 0 \\
   -1 & x \leq 0.
  \end{cases}
\end{equation}

\subsection{Support Vector Machines}
Support Vector Machines (SVM) involves determining the boundary that maximizes the margin between the data based on a hinge loss $L_{hinge}(w,b)=max(0,1-y f(x))$, where $yf(x)$ gives the margin \cite{vapnik}. This translates to finding the optimal model parameters $(w^*,b^*)$ by solving the following optimization problem
\begin{equation}\label{primal objective}
\min_{w,b,\zeta_i} \frac{1}{2}w^Tw+C \sum_{i=1}^{N}{\zeta_i}
\end{equation}
\begin{equation}\label{cons}
 \text{subjected to}
  \begin{cases}
   y_i[\langle w,\phi(x_i)\rangle+b]\geq 1-\zeta_i \\
    \zeta_i \geq 0
  \end{cases}\\
\end{equation}
for $i=1,..,N$. Here $\zeta_i$ represents the slack variable for data observation $i$, $C$ represents the cost penalty hyper-parameter. The slack variables $\zeta_i$ are required in order to allow for misclassifications in a noisy overlapping binary data set that cannot be completely separated by a decision boundary. The input vectors $x$ are mapped onto a higher dimensional space using the function $\phi$. By making this transformation, the nonlinear data is aligned linearly in the high dimensional space where SVM finds a maximum margin separating hyperplane. The transformation is performed implicitly using a kernel matrix $K(x_i,x_j)=[k(x_i,x_j)]_{i,j}$ where $k(x_i,x_j)$ could be any function satisfying Mercer's condition \cite{vapnik}. The gaussian kernel function (equation \eqref{ker_fn}) is used in this paper. More details on SVM formulation can be found in \cite{vapnik}.
\begin{equation}\label{ker_fn}
k(x_i,x_j)=e^{-\sigma\|x_i-x_j\|^2}, \sigma >0
\end{equation}

The convex constrained optimization problem in equation \eqref{primal objective} is in the primal form, and the variables $w, b$ and $\zeta_i$ are referred to as primal variables. The primal problem is converted to a dual formulation in equation \eqref{dualeqn} and solved for the dual variables $\alpha_i$.
\begin{equation}\label{dualeqn}
    \max_{\alpha_i} \left\{-\frac{1}{2}\sum_{i=1}^N{\sum_{j=1}^N{y_i y_j\alpha_i \alpha_j K(x_i,x_j)}}+\sum_{i=1}^N{\alpha_i}\right\}
\end{equation}
\begin{equation}\label{}
 \text{subjected to}
  \begin{cases}
      \sum_{i=1}^N{\alpha_i y_i}=0\\
      0\leq\alpha_i\leq C\\
  \end{cases}
\end{equation}
for $i=1,..,N$. The SVM hypothesis is given by
\begin{equation}\label{SVR model_dual}
f(x)=sgn \left(\sum_{i=1}^N{\alpha_i y_i K(x_i,x)}+b \right)
\end{equation}
The above formulation is not designed for an imbalanced data set where the majority class data outnumbers the minority class data. A cost-sensitive version of the SVM algorithm is used in such cases, where the cost penalty parameter $C$ in equation \eqref{primal objective} is modified to weigh more to the penalties of the minority class data compared to the majority class data \cite{libsvm,svm_imbalanced}. All implementations of SVM are done using LibSVM \cite{libsvm}. The cost modification can be performed as follows
\begin{equation}\label{weight_fac_svm}
 C_i =
  \begin{cases}
   C &  \text{majority class data} \\
   C(r\times f)  & \text{minority class data}
  \end{cases}
\end{equation}
where $r$ represents the ratio of number of majority class data to the number of minority class data and $f$ represents a scaling factor to be tuned for a given data set.

\subsection{Extreme Learning Machines}
Extreme Learning Machine is an emerging learning paradigm for multi-class classification and regression problems \cite{4Huang2005,huang12}. The highlight of ELM is that the training speed is extremely fast (or computationally inexpensive). The key enabler for ELM's training speed is the random assignment of input layer parameters which do not require adaptation to the data. In such a setup, the output layer parameters can be determined analytically using a least squares approach. Some of the attractive features of ELM \cite{4Huang2005} include the universal approximation capability of ELM, the convex optimization problem of ELM resulting in the smallest training error without getting trapped in local minima, closed form solution of ELM eliminating iterative training and better generalization capability of ELM. ELM training involves solving the following optimization problem
\begin{equation}\label{ELM_opti}
\min_{W}\left\{\|HW-Y\|^2+\lambda \|W\|^2\right\}
\end{equation}
\begin{equation}\label{}
H^T=\psi(W_r^Tx(k)+b_r)\in\mathbb{R}^{n_h \times 1}
\end{equation}
where $\lambda$ represents the regularization coefficient, Y represents the vector of outputs or targets, $\psi$ represents the hidden layer activation function (a sigmoidal function takes the same structure as \eqref{logistic}) and $W_r\in\mathbb{R}^{n \times n_h}, W\in\mathbb{R}^{n_h \times 1}$ represents the input and output layer parameters respectively. Here, $n_h$ represents the number of hidden neurons of the ELM model, $H$ represents the hidden layer output matrix. The matrix $W_r$ consists of randomly assigned elements that maps the input vector to a high dimensional feature space while $b_r\in\mathbb{R}^{n_h}$ is a bias component assigned in a random manner similar to $W_r$. The number of hidden neurons determines the dimension of the transformed feature space. The elements can be assigned based on any continuous random distribution \cite{huang12} and remains fixed during the learning process. Hence the training reduces to a single step calculation given by equation \eqref{ELM_train}. The ELM decision hypothesis can be expressed as in equation \eqref{elm model}. It should be noted that the hidden layer and the corresponding activation functions give a nonlinear mapping of the data, which if eliminated, becomes a linear least squares (Linear LS) model and is considered as one of the baseline models in this study.
\begin{equation}\label{ELM_train}
W^*=\left(H^TH + \lambda I \right)^{-1}H^TY
\end{equation}
\begin{equation}\label{elm model}
f(x)=sgn\left(W^T[\psi(W_r^Tx+b_r)]\right)
\end{equation}

The above ELM formulation is not designed to handle imbalanced or skewed data sets. As a modification to weigh the minority class data more, a simple weighting method can be incorporated in the ELM objective function \eqref{ELM_opti} as
\begin{equation}\label{ELM_opti_weight}
\min_{W}\left\{(HW-Y)^T\Gamma(HW-Y)+\lambda W^T W\right\}
\end{equation}
\begin{center}
$\Gamma = \left [ \begin{array}{ccccc}
\gamma_1 & 0 & .& .& 0\\
0 & \gamma_2 & .& .& 0\\
. & . & .& .& 0\\
0 & 0 & .& .& \gamma_N\\
\end{array} \right ]$
\end{center}
\begin{equation}\label{weight_fac}
 \gamma_i =
  \begin{cases}
   1 &  \text{majority class data} \\
   r \times f  & \text{minority class data}
  \end{cases}
\end{equation}
where $\Gamma$ represents the weight matrix, $r$ represents the ratio of number of majority class data to number minority class data and $f$ represents a scaling factor to be tuned for a given data set. This results in the training step given by equation \eqref{ELM_train_weight} and hypothesis given by equation \eqref{elm model_weight}.
\begin{equation}\label{ELM_train_weight}
W^*=\left(H^T \Gamma H + \lambda I \right)^{-1}H^T \Gamma Y
\end{equation}
\begin{equation}\label{elm model_weight}
f(x)=sgn\left(W^T[\psi(W_r^Tx+b_r)]\right)
\end{equation}

\section{HCCI engine and data processing} \label{system}
For the purpose of identifying the stable operating envelope of HCCI engine, transient experiments are performed by exciting the engine and recording time sequences of engine variables. In this section, the HCCI engine system and experiments performed are briefly explained followed by a methodology of labeling the data suitable for classification.

\subsection{HCCI System and Experimentation}\label{sec:expt}
\begin{figure*}[hptb]
      \centering
      \includegraphics[scale=0.5]{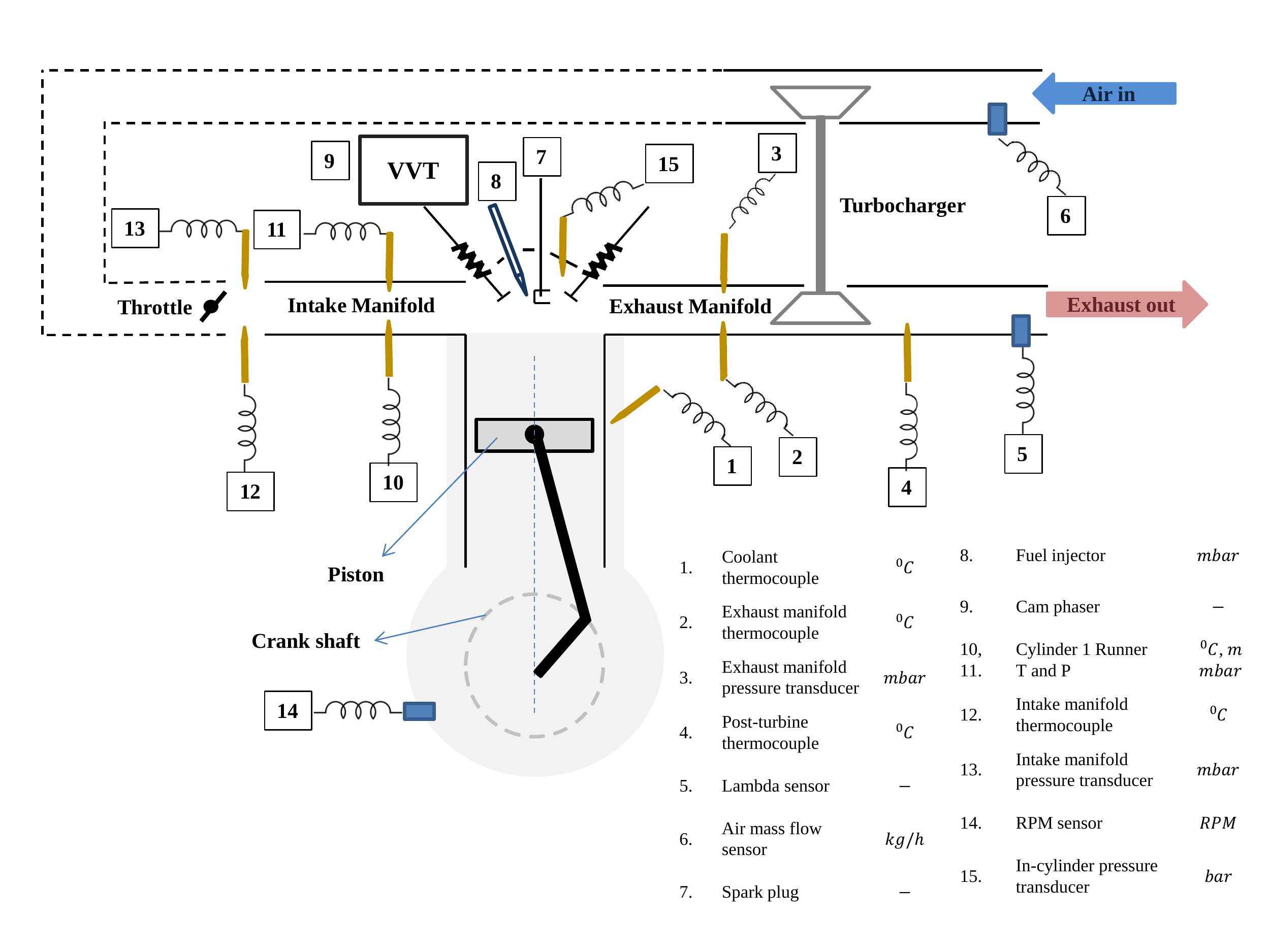}
      \caption{A schematic of the HCCI engine setup and instrumentation (only relevant instrumentation is shown).}
      \label{schematic}
\end{figure*}

The concerned system of interest is a gasoline HCCI engine with a variable valve timing system. The engine specifications are listed in Table \cite{vijay_asoc}. A schematic of the experimental setup and instrumentation is shown in Fig. \ref{schematic}. HCCI is achieved by auto-ignition of the gas mixture in the cylinder. The fuel is injected early and given sufficient time to mix with air forming a homogeneous mixture. A large fraction of exhaust gas from the previous combustion cycle is retained to elevate the temperature and hence the reaction rates of the fuel and air mixture. The variable valve timing capability of the engine enables trapping suitable quantities of exhaust gas in the cylinder.

The engine can be controlled using precalculated inputs such as injected fuel mass (FM in mg/cyc), crank angle at intake valve opening (IVO), crank angle at exhaust valve closing (EVC), crank angle at start of fuel injection (SOI). Other important physical variables that influence the performance of HCCI combustion include intake manifold temperature $T_{in}$, intake manifold pressure $P_{in}$, mass flow rate of air at intake $\dot{m}_{in}$, exhaust gas temperature $T_{ex}$, exhaust manifold pressure $P_{ex}$, coolant temperature $T_{c}$, fuel to air ratio (FA) etc. The engine performance metrics are given by combustion phasing indicated by the crank angle at 50\% mass fraction burned (CA50), combustion work output indicated by net mean effective pressure (NMEP). Both CA50 and NMEP are determined from the high speed in-cylinder pressure measurements. The above variables at present time instant $k$ along with their time histories are considered as inputs to the model (see section \ref{modeling}, equation \eqref{app_inp}). For further reading on HCCI combustion and related variables, please refer \cite{hcci_book}.

As mentioned in section \ref{intro}, the goal of this paper is to identify the HCCI operating boundary in transient operation. This requires an appropriate experiment design to obtain transient data from the engine. A set of transient experiments is conducted at a constant speed of 2500 RPM and naturally aspirated conditions by varying FM, IVO, EVC and SOI in a uniformly random manner. The experiments are conducted and data recorded using specialized engine rapid prototyping hardware. An amplitude modulated pseudo-random binary sequence (A-PRBS) has been used to design the excitation signals. A-PRBS enables exciting the engine at different amplitudes and frequencies suitable for the identification problem considered in this work. The data is sampled using the AVL Indiset acquisition system where in-cylinder pressure is sensed every crank angle while NMEP, CA50 and $R_{max}$ are determined on a per-combustion cycle basis. More details on HCCI combustion and experiments can be found in \cite{vijay_springer}. The data is pre-processed and labeled to identify stable and unstable observations as explained in section \ref{sec:labeling}.

\begin{table}[]
\caption{Specifications of the experimental HCCI engine}
\label{specstable}
\footnotesize
\begin{center}
\begin{tabular}[c]{|c|c|}
\hline
Engine Type & 4-stroke In-line\\
\hline
Fuel & Gasoline\\
\hline
Displacement & 2.0 L\\
\hline
Bore/Stroke & 86/86 mm\\
\hline
Compression Ratio & 11.25:1\\
\hline
Injection Type & Direct Injection\\
\hline
 & Variable Valve Timing with \\
&  hydraulic cam phaser having\\
Valvetrain  & 119 degree constant duration \\
 & defined at 0.25mm lift, 3.5mm peak \\
 & lift and 50 degree crank angle \\
 & phasing authority \\
 \hline
HCCI strategy & Exhaust recompression \\
& using negative valve overlap\\
\hline
\end{tabular}
\end{center}
\end{table}

\subsection{HCCI Instabilities}
\begin{figure}[]
      \centering
      \includegraphics[scale=0.35]{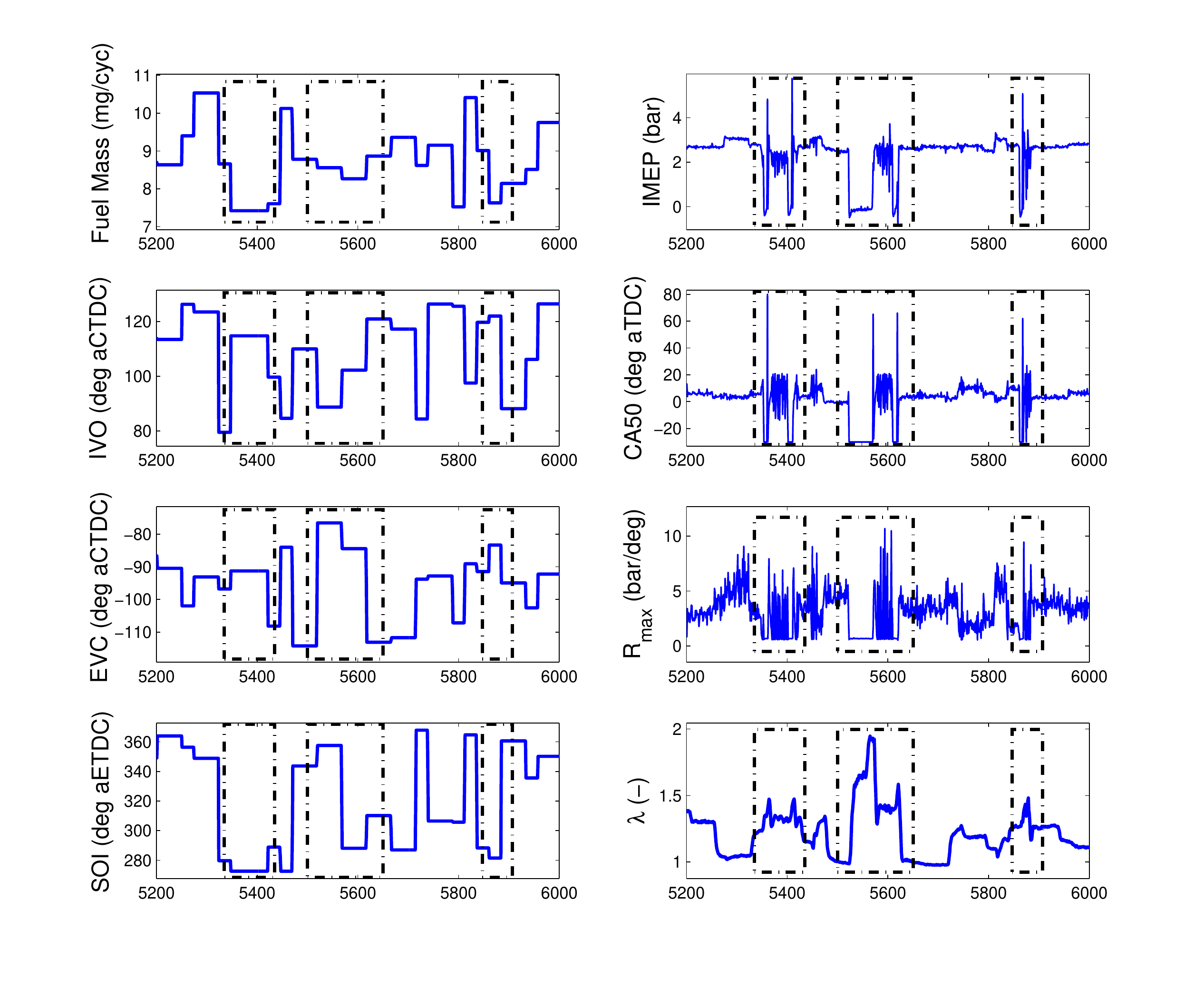}
      \caption{A-PRBS inputs and outputs showing misfire regions.}
      \label{misfire fig}
\end{figure}

A subset of the data collected from the engine is shown in Figure \ref{misfire fig} where it can be observed that for some combinations of the inputs (left figures), the HCCI engine misfires (seen in the right figures where IMEP drops below 0 bar). HCCI operation is limited by several phenomena that lead to undesirable engine behavior. As described in \cite{george}, the HCCI operating range is conceptually constrained to a small region of permissible unburned (pre-combustion) and burned (post-combustion) charge temperature states. As previously noted, sufficiently high unburned gas temperatures are required to achieve ignition in the HCCI operating range without which complete misfire will occur. If the resulting combustion cannot achieve sufficiently high burned gas temperatures, commonly occurring in conditions with low fuel to diluent ratios or late combustion phasing, various degrees of quenching can occur resulting in reduced work output and increased hydrocarbon and carbon monoxide emissions. Under some conditions, this may lead to high cyclic variation due to the positive feedback loop existing through the trapped residual gas
\cite{misf1,misf2}. Operation with high burned gas temperature, although stable and commonly reached at higher fueling rates where the fuel to diluent ratio is also high, yields high heat release and thus pressure rise rates that may pose challenges for engine noise and durability constraints. A discussion of the temperatures at which these phenomena occur may be found in \cite{george}.

\begin{figure}[]
      \centering
      \includegraphics[scale=0.47]{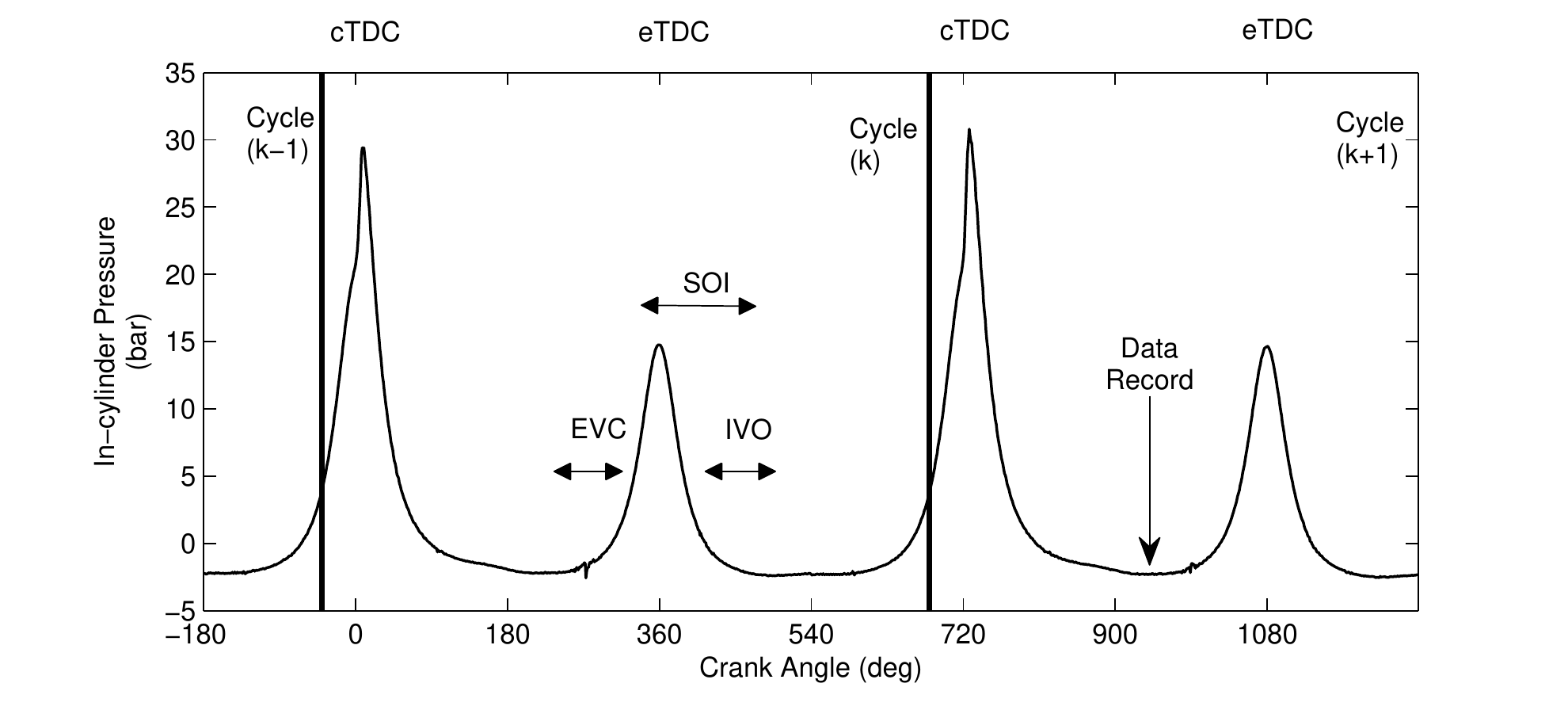}
      \caption{HCCI combustion cycle definition showing valve events (IVO and EVC), injected fuel mass (FM) and start of injection (SOI). Negative valve overlap can be seen as a smaller peak at eTDC.}
      \label{cyc_defn}
\end{figure}

In this paper, the considered instabilities include those modes with high cyclic variability and those with complete misfire characterized by zero work output that can be readily identified through the two aforementioned cylinder pressure-based combustion features. The other phenomena could be included with the availability of additional sensing capability or analysis methods, e.g. fast response Flame Ionization Detection exhaust sampling equipment and detailed combustion noise analysis. Finally, it must be noted that control of these burned and unburned gas states, and therefore the potential for undesirable combustion cycles, in a recompression HCCI engine is very much a function of the engine control variables. For instance, the EVC timing will determine the trapped residual mass that will be present in the upcoming cycle, while the IVO affects both the mass of incoming air and the state of the charge during the compression stroke leading up to the autoignition. The combination of IVO and EVC (see Fig. \ref{cyc_defn}) define a negative valve overlap (NVO) period where exhaust gas from the previous cycle is trapped and compressed. A larger NVO period would necessarily yield a higher trapped residual mass that would tend to increase the charge temperature and advance CA50. Likewise, the timing and mass of the fuel injection event can significantly impact the charge temperature by changing the thermodynamic properties and air-fuel ratio of the charge present during NVO. The relatively high temperatures present during NVO can even lead to reactions of the fuel that will impact the temperature and chemical composition of the charge. Successful combustion of charge with a higher FM will tend to yield higher residual gas temperatures, thereby advancing CA50 in the following cycle. Likewise an earlier SOI in NVO will tend to increase charge temperatures and reduce the ignition delay of the charge, thereby advancing CA50. As such, an improper combination of control inputs (IVO, EVC, FM and SOI) in HCCI engines has the potential to shift operation from stable combustion  to combustion with excessive heat release rates, high cyclic variability or misfires in a single cycle.

\subsection{Data Preprocessing and Labeling} \label{sec:labeling}
\begin{figure}
      \centering
      \includegraphics[scale=0.33]{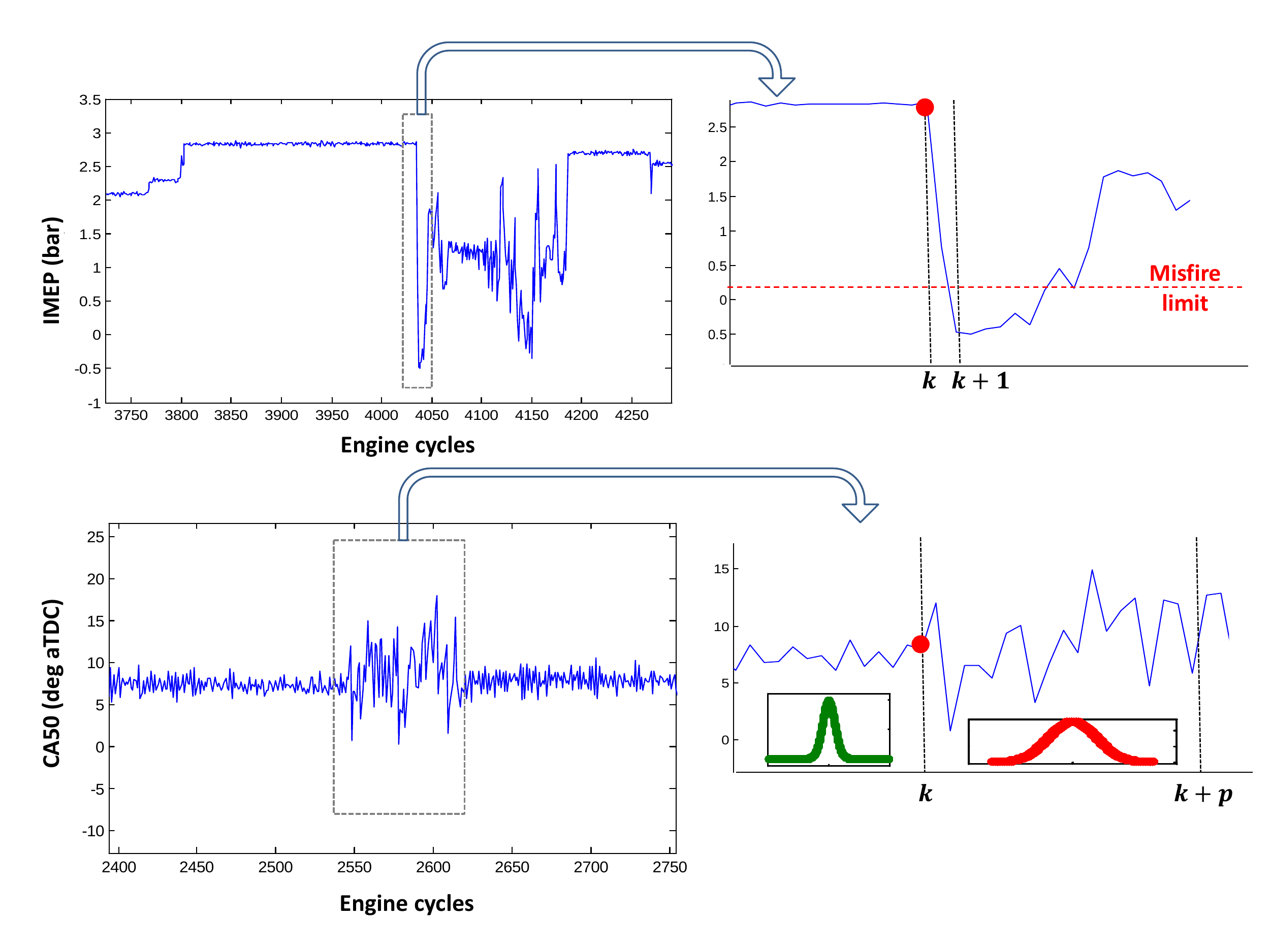}
      \caption{Illustration showing labeling of unstable observations.}
      \label{labeling_unst}
\end{figure}

\begin{figure}
      \centering
      \includegraphics[scale=0.33]{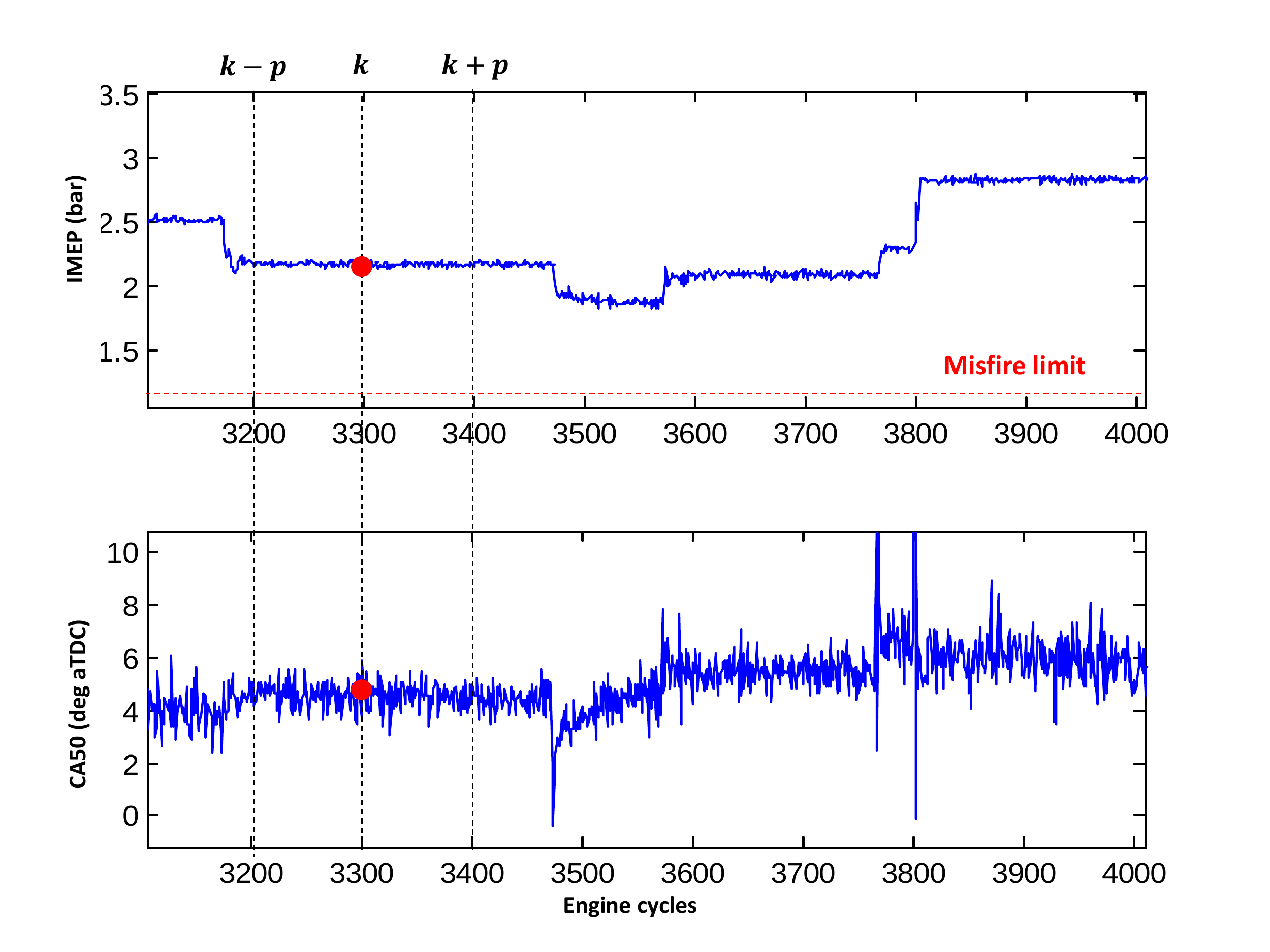}
      \caption{Illustration showing labeling of stable observations.}
      \label{labeling_st}
\end{figure}

The goal of the learning algorithm is to classify the input space into stable (future HCCI cycles are stable) or unstable (future HCCI cycles misfire or have high variability) engine behavior. The input space includes sensor measurements until the present time instant while the indicator label depends on the future. For this purpose, the data is labeled as follows. If either of the following two conditions are met, then the data at time instant $k$ is labeled to be unstable (see Fig. \ref{labeling_unst}) with a label value -1.
\begin{enumerate}
  \item an input (control inputs and past engine measurements up to an order of $N_h$) at cycle $k$ results in an IMEP of less than 0.1 bar (chosen misfire limit) for any cylinder at cycle $k+1$.
  \item an input at cycle $k$ results in a high variance of CA50 (any cylinder) for cycles $k+1$ to $k+p$.
\end{enumerate}
Only the first unstable data is considered in a sequence of unstable measurements. The labeling of stable data is as follows. A window of $N_w=2p$ combustion cycles is considered (see Fig. \ref{labeling_st}). If the data at cycle $k$ is obtained as a result of stable operation in the past $p$ cycles as well as results in stable operation in the next $p$ cycles, it is labeled stable with a label value of +1. If the time history of $N_h$ is considered, then the data at cycle $k$ along with the previous $N_h$ samples are considered as inputs. If $N_h$ is a large value, then the window length $N_w$ can be increased accordingly. In this study, $N_w$ and $N_h$ correspond to 10 and 2 respectively.

\section{Modeling the Stable Operating Envelope of HCCI Engine} \label{modeling}
The HCCI operating envelope is a function of the engine control inputs and engine variables such as temperatures and pressures. Also, the envelope is a dynamic system and hence a predictive model requires the measurement history up to an order of $N_h$. The dynamic classifier model can be given by
\begin{equation}\label{}
\hat{y}_{k+1}=sgn(f(x_k))
\end{equation}
where $\hat{y}_{k+1}$ indicates model prediction for the future cycle $k+1$, $f$ can take any structure depending on the learning algorithm and $x_k$ is given by
\begin{multline}\label{app_inp}
x_k=[IVO, EVC, FM, SOI, T_{in}, P_{in}, \dot{m}_{in}, \\
    T_{ex}, P_{ex}, T_{c}, FA, IMEP, CA50]^T
\end{multline}
at cycle $k$ upto cycle $k-N_h+1$.

\subsection{Model Selection}
In this section, classification algorithms are developed based on Linear Regression (LS), Logistic regression (LR), SVM and ELM models. SVM and ELM models have variants based on under-sampling, over-sampling or no-sampling (regular) the data set or the cost-sensitive version. The linear models (LR and LS) are compared as baselines and have their respective variants. The engine measurements and their time histories (defined by $x_k$) are considered inputs while the stability labels are considered outputs. The measured variables such as FM, IVO, EVC, SOI, $T_c$, $T_{in}$, $P_{in}$, $\dot{m}_{in}$, $T_{ex}$, $P_{ex}$, NMEP, CA50 and FA along with 2 cycles of history constitute the feature vector (input dimension $n$ = 39). The measurement set consists of about 17000 observations out of which about 6400 observations are sampled as training set while about 10200 observations sampled as testing set. The ratio of number of majority class data to number minority class data ($r$) for the training set is 17.5 and for the testing set is 16.7.

For the class imbalance problem considered here, a typical error metric like the overall misclassification rate cannot be used as it would find a classifier that ignores the minor class inaccuracies. Hence the following accuracy metric for skewed data sets are considered. Let $TP$ and $TN$ represent the total number of positive and negative class data classified correctly by the classifier. If $N^+$ and $N^-$ represent the total number of positive and negative class data respectively, the true positive rate (TPR) and true negative rate (TNR) and the total accuracy of the classifier can be defined as follows \cite{toh_imbalanced}

\begin{eqnarray}
\nonumber TPR &=& \frac{TP}{N^+} \\
\nonumber TNR &=& \frac{TN}{N^-} \\
\text{Total Accuracy} &=& \frac{TPR+TNR}{2}
\end{eqnarray}

\begin{table*}[]
  \centering
  \caption{Grid search results for ELM model selection for the regular ELM, ELM with under-sampling and ELM with over-sampling (The models resulting in lowest total accuracy is highlighted in bold).}
  \footnotesize
    \begin{tabular}{cccccccccccccccccc}
    \hline
              &       &       &       &       &       &       &       &       &       &       &       &       &       &       &       &       &  \\
        & \multicolumn{5}{c}{Regular ELM}               &       & \multicolumn{5}{c}{ELM with under-sampling}               &       & \multicolumn{5}{c}{ELM with over-sampling} \\
                  &       &       &       &       &       &       &       &       &       &       &       &       &       &       &       &       &  \\
    \hline
          &       &       &       &       &       &       &       &       &       &       &       &       &       &       &       &       &  \\
    & \multicolumn{5}{c}{TPR}               &       & \multicolumn{5}{c}{TPR}               &       & \multicolumn{5}{c}{TPR} \\
       \backslashbox{$n_h$}{$\lambda$}   & 0.01  & 0.1   & 1     & 10    & 100   &       & 0.01  & 0.1   & 1     & 10    & 100   &       & 0.01  & 0.1   & 1     & 10    & 100 \\
    10    & 0.995 & 0.995 & 0.995 & 0.996 & 1.000 &       & 0.909 & 0.909 & 0.912 & 0.906 & 0.878 &       & 0.925 & 0.925 & 0.925 & 0.925 & 0.919 \\
    30    & 0.994 & 0.994 & 0.995 & 0.995 & 0.997 &       & 0.917 & 0.916 & 0.918 & 0.923 & 0.896 &       & 0.934 & 0.934 & 0.934 & 0.936 & 0.936 \\
    50    & 0.995 & 0.995 & 0.995 & 0.995 & 0.998 &       & 0.917 & 0.918 & 0.925 & 0.948 & 0.915 &       & 0.930 & 0.931 & 0.932 & 0.935 & 0.951 \\
    70    & 0.996 & 0.996 & 0.996 & 0.996 & 0.996 &       & 0.936 & 0.937 & 0.931 & 0.936 & 0.948 &       & 0.944 & 0.945 & 0.946 & 0.946 & 0.946 \\
    90    & \textbf{0.995} & 0.995 & 0.995 & 0.995 & 0.996 &       & 0.915 & \textbf{0.918} & 0.924 & 0.929 & 0.927 &       & 0.938 & 0.938 & 0.939 & \textbf{0.944} & 0.942 \\
          &       &       &       &       &       &       &       &       &       &       &       &       &       &       &       &       &  \\
          & \multicolumn{5}{c}{TNR}               &       & \multicolumn{5}{c}{TNR}               &       & \multicolumn{5}{c}{TNR} \\
       \backslashbox{$n_h$}{$\lambda$}   & 0.01  & 0.1   & 1     & 10    & 100   &       & 0.01  & 0.1   & 1     & 10    & 100   &       & 0.01  & 0.1   & 1     & 10    & 100 \\
    10    & 0.333 & 0.333 & 0.327 & 0.258 & 0.000 &       & 0.732 & 0.734 & 0.737 & 0.771 & 0.743 &       & 0.716 & 0.716 & 0.716 & 0.722 & 0.757 \\
    30    & 0.387 & 0.389 & 0.387 & 0.366 & 0.160 &       & 0.714 & 0.719 & 0.730 & 0.732 & 0.732 &       & 0.727 & 0.725 & 0.729 & 0.735 & 0.727 \\
    50    & 0.430 & 0.430 & 0.423 & 0.407 & 0.294 &       & 0.771 & 0.773 & 0.752 & 0.735 & 0.727 &       & 0.773 & 0.773 & 0.771 & 0.755 & 0.724 \\
    70    & 0.426 & 0.423 & 0.413 & 0.404 & 0.351 &       & 0.771 & 0.770 & 0.775 & 0.748 & 0.704 &       & 0.739 & 0.739 & 0.739 & 0.758 & 0.740 \\
    90    & \textbf{0.433} & 0.430 & 0.420 & 0.405 & 0.356 &       & 0.789 & \textbf{0.794} & 0.784 & 0.768 & 0.755 &       & 0.763 & 0.768 & 0.773 & \textbf{0.775} & 0.775 \\
          &       &       &       &       &       &       &       &       &       &       &       &       &       &       &       &       &  \\
          & \multicolumn{5}{c}{Total Accuracy}    &       & \multicolumn{5}{c}{Total Accuracy}    &       & \multicolumn{5}{c}{Total Accuracy} \\
       \backslashbox{$n_h$}{$\lambda$}   & 0.01  & 0.1   & 1     & 10    & 100   &       & 0.01  & 0.1   & 1     & 10    & 100   &       & 0.01  & 0.1   & 1     & 10    & 100 \\
    10    & 0.664 & 0.664 & 0.661 & 0.627 & 0.500 &       & 0.821 & 0.821 & 0.825 & 0.839 & 0.811 &       & 0.820 & 0.820 & 0.820 & 0.824 & 0.838 \\
    30    & 0.691 & 0.692 & 0.691 & 0.680 & 0.579 &       & 0.815 & 0.818 & 0.824 & 0.828 & 0.814 &       & 0.831 & 0.830 & 0.832 & 0.835 & 0.832 \\
    50    & 0.712 & 0.712 & 0.709 & 0.701 & 0.646 &       & 0.844 & 0.846 & 0.838 & 0.842 & 0.821 &       & 0.852 & 0.852 & 0.851 & 0.845 & 0.838 \\
    70    & 0.711 & 0.709 & 0.705 & 0.700 & 0.674 &       & 0.853 & 0.853 & 0.853 & 0.842 & 0.826 &       & 0.842 & 0.842 & 0.842 & 0.852 & 0.843 \\
    90    & \textbf{0.714} & 0.712 & 0.707 & 0.700 & 0.676 &       & 0.852 & \textbf{0.856} & 0.854 & 0.848 & 0.841 &       & 0.850 & 0.853 & 0.856 & \textbf{0.859} & 0.858 \\
          &       &       &       &       &       &       &       &       &       &       &       &       &       &       &       &       &  \\
    \hline
    \end{tabular}%
  \label{ELM_tuning1}%
\end{table*}%
\begin{table*}[]
  \centering
  \footnotesize
  \caption{Grid search results for SVM model selection for the regular SVM, SVM with under-sampling and SVM with over-sampling (The models resulting in lowest total accuracy is highlighted in bold).}
    \begin{tabular}{cccccccccccccccccc}
    \hline
              &       &       &       &       &       &       &       &       &       &       &       &       &       &       &       &       &  \\
    \multicolumn{6}{c}{Regular SVM}                               &       & \multicolumn{5}{c}{SVM with under-sampling}                       &       & \multicolumn{5}{c}{SVM with over-sampling} \\
              &       &       &       &       &       &       &       &       &       &       &       &       &       &       &       &       &  \\
    \hline
              &       &       &       &       &       &       &       &       &       &       &       &       &       &       &       &       &  \\
    \multicolumn{6}{c}{TPR}                       &       & \multicolumn{5}{c}{TPR}               &       & \multicolumn{5}{c}{TPR} \\
    \backslashbox{$C$}{$\sigma$} & 0.01  & 0.1   & 1     & 10    & 100   &       & 0.01  & 0.1   & 1     & 10    & 100   &       & 0.01  & 0.1   & 1     & 10    & 100 \\
    0.1   & 1.000 & 0.996 & 1.000 & 1.000 & 1.000 &       & 0.928 & 0.933 & 0.767 & 0.120 & 0.106 &       & 0.966 & 0.923 & 0.917 & \textbf{0.899} & 0.993 \\
    1     & 0.996 & 0.996 & 0.994 & 0.998 & 1.000 &       & 0.965 & 0.906 & 0.910 & 0.782 & 0.588 &       & 0.932 & 0.931 & 0.962 & 0.989 & 0.999 \\
    10    & 0.996 & 0.995 & 0.990 & 0.997 & 0.999 &       & 0.916 & 0.915 & \textbf{0.909} & 0.792 & 0.618 &       & 0.933 & 0.951 & 0.976 & 0.996 & 0.999 \\
    100   & 0.996 & 0.990 & \textbf{0.987} & 0.996 & 0.999 &       & 0.924 & 0.927 & 0.896 & 0.793 & 0.618 &       & 0.935 & 0.966 & 0.983 & 0.996 & 0.999 \\
    500   & 0.995 & 0.988 & 0.985 & 0.996 & 0.999 &       & 0.925 & 0.917 & 0.892 & 0.793 & 0.618 &       & 0.945 & 0.971 & 0.983 & 0.996 & 0.999 \\
          &       &       &       &       &       &       &       &       &       &       &       &       &       &       &       &       &  \\
    \multicolumn{6}{c}{TNR}                       &       & \multicolumn{5}{c}{TNR}               &       & \multicolumn{5}{c}{TNR} \\
       \backslashbox{$C$}{$\sigma$}   & 0.01  & 0.1   & 1     & 10    & 100   &       & 0.01  & 0.1   & 1     & 10    & 100   &       & 0.01  & 0.1   & 1     & 10    & 100 \\
    0.1   & 0.000 & 0.374 & 0.000 & 0.000 & 0.000 &       & 0.632 & 0.735 & 0.931 & 0.998 & 0.998 &       & 0.667 & 0.820 & 0.913 & \textbf{0.958} & 0.162 \\
    1     & 0.397 & 0.423 & 0.552 & 0.108 & 0.054 &       & 0.668 & 0.825 & 0.923 & 0.967 & 0.987 &       & 0.792 & 0.884 & 0.814 & 0.221 & 0.082 \\
    10    & 0.423 & 0.444 & 0.645 & 0.145 & 0.080 &       & 0.802 & 0.882 & \textbf{0.925} & 0.958 & 0.975 &       & 0.822 & 0.817 & 0.732 & 0.167 & 0.080 \\
    100   & 0.430 & 0.567 & \textbf{0.650} & 0.165 & 0.080 &       & 0.833 & 0.886 & 0.915 & 0.954 & 0.975 &       & 0.848 & 0.763 & 0.642 & 0.165 & 0.080 \\
    500   & 0.436 & 0.627 & 0.637 & 0.165 & 0.080 &       & 0.853 & 0.882 & 0.910 & 0.954 & 0.975 &       & 0.833 & 0.724 & 0.623 & 0.165 & 0.080 \\
          &       &       &       &       &       &       &       &       &       &       &       &       &       &       &       &       &  \\
    \multicolumn{6}{c}{Total Accuracy}            &       & \multicolumn{5}{c}{Total Accuracy}    &       & \multicolumn{5}{c}{Total Accuracy} \\
      \backslashbox{$C$}{$\sigma$}    & 0.01  & 0.1   & 1     & 10    & 100   &       & 0.01  & 0.1   & 1     & 10    & 100   &       & 0.01  & 0.1   & 1     & 10    & 100 \\
    0.1   & 0.500 & 0.685 & 0.500 & 0.500 & 0.500 &       & 0.780 & 0.834 & 0.849 & 0.559 & 0.552 &       & 0.816 & 0.871 & 0.915 & \textbf{0.928} & 0.577 \\
    1     & 0.696 & 0.710 & 0.773 & 0.553 & 0.527 &       & 0.817 & 0.866 & 0.916 & 0.875 & 0.787 &       & 0.862 & 0.907 & 0.888 & 0.605 & 0.540 \\
    10    & 0.709 & 0.720 & 0.818 & 0.571 & 0.539 &       & 0.859 & 0.899 & \textbf{0.917} & 0.875 & 0.797 &       & 0.878 & 0.884 & 0.854 & 0.581 & 0.539 \\
    100   & 0.713 & 0.779 & \textbf{0.819} & 0.580 & 0.539 &       & 0.879 & 0.906 & 0.905 & 0.874 & 0.797 &       & 0.892 & 0.865 & 0.812 & 0.580 & 0.539 \\
    500   & 0.716 & 0.808 & 0.811 & 0.580 & 0.539 &       & 0.889 & 0.899 & 0.901 & 0.874 & 0.797 &       & 0.889 & 0.847 & 0.803 & 0.580 & 0.539 \\
              &       &       &       &       &       &       &       &       &       &       &       &       &       &       &       &       &  \\
\hline
    \end{tabular}%
  \label{SVM_tuning1}%
\end{table*}%

\begin{table*}[]
\footnotesize
  \centering
  \caption{Grid search results for Cost-sensitive SVM and Cost-sensitive ELM models (The models resulting in lowest total accuracy is highlighted in bold).}
    \begin{tabular}{cccccccccccccc}
    \hline
          &       &       &       &       &       &       &       &       &       &       &       &       &  \\
    \multicolumn{7}{c}{Cost-sensitive SVM}                               &       &       & \multicolumn{5}{c}{Cost-sensitive ELM} \\
      &       &       &       &       &       &       &       &       &       &       &       &       &  \\
          \hline
          &       &       &       &       &       &       &       &       &       &       &       &       &  \\
    \multicolumn{7}{c}{TPR}                               &       &       & \multicolumn{5}{c}{TPR} \\
    \backslashbox{$C$}{$\sigma$} & 0.001 & 0.01  & 0.1   & 1     & 10    & 100   &       &   \backslashbox{$n_h$}{$\lambda$}    & 0.01  & 0.1   & 1     & 10    & 100 \\
    0.1   & 0.000 & 0.000 & 0.000 & 0.000 & 0.000 & 0.000 &       & 10    & 0.909 & 0.909 & 0.909 & 0.907 & 0.901 \\
    1     & 0.984 & 0.972 & 0.921 & 0.918 & \textbf{0.884} & 0.764 &       & 30    & 0.924 & 0.924 & 0.923 & 0.927 & 0.927 \\
    10    & 0.995 & 0.994 & 0.993 & 0.988 & 0.996 & 0.999 &       & 50    & 0.925 & 0.925 & 0.926 & 0.930 & 0.936 \\
    100   & 1.000 & 0.999 & 0.993 & 0.987 & 0.996 & 0.999 &       & 70    & 0.939 & 0.939 & 0.939 & 0.934 & 0.931 \\
    500   & 1.000 & 0.997 & 0.989 & 0.985 & 0.996 & 0.999 &       & 90    & 0.932 & 0.932 & \textbf{0.933} & 0.936 & 0.930 \\
          &       &       &       &       &       &       &       &       &       &       &       &       &  \\
    \multicolumn{7}{c}{TNR}                               &       &       & \multicolumn{5}{c}{TNR} \\
          & 0.001 & 0.01  & 0.1   & 1     & 10    & 100   &       &       & 0.01  & 0.1   & 1     & 10    & 100 \\
    0.1   & 1.000 & 1.000 & 1.000 & 1.000 & 1.000 & 1.000 &       & 10    & 0.742 & 0.743 & 0.745 & 0.753 & 0.779 \\
    1     & 0.490 & 0.627 & 0.806 & 0.912 & \textbf{0.961} & 0.972 &       & 30    & 0.737 & 0.737 & 0.735 & 0.740 & 0.735 \\
    10    & 0.399 & 0.443 & 0.489 & 0.660 & 0.154 & 0.080 &       & 50    & 0.778 & 0.778 & 0.775 & 0.771 & 0.748 \\
    100   & 0.181 & 0.355 & 0.495 & 0.647 & 0.165 & 0.080 &       & 70    & 0.770 & 0.770 & 0.765 & 0.779 & 0.768 \\
    500   & 0.126 & 0.405 & 0.554 & 0.639 & 0.165 & 0.080 &       & 90    & 0.786 & 0.786 & \textbf{0.794} & 0.784 & 0.784 \\
          &       &       &       &       &       &       &       &       &       &       &       &       &  \\
    \multicolumn{7}{c}{Total Accuracy}                    &       &       & \multicolumn{5}{c}{Total Accuracy} \\
          & 0.001 & 0.01  & 0.1   & 1     & 10    & 100   &       &       & 0.01  & 0.1   & 1     & 10    & 100 \\
    0.1   & 0.500 & 0.500 & 0.500 & 0.500 & 0.500 & 0.500 &       & 10    & 0.825 & 0.826 & 0.827 & 0.830 & 0.840 \\
    1     & 0.737 & 0.800 & 0.863 & 0.915 & \textbf{0.922} & 0.868 &       & 30    & 0.830 & 0.830 & 0.829 & 0.833 & 0.831 \\
    10    & 0.697 & 0.718 & 0.741 & 0.824 & 0.575 & 0.539 &       & 50    & 0.852 & 0.851 & 0.850 & 0.851 & 0.842 \\
    100   & 0.591 & 0.677 & 0.744 & 0.817 & 0.580 & 0.539 &       & 70    & 0.854 & 0.854 & 0.852 & 0.857 & 0.850 \\
    500   & 0.563 & 0.701 & 0.771 & 0.812 & 0.580 & 0.539 &       & 90    & 0.859 & 0.859 & \textbf{0.864} & 0.860 & 0.857 \\
          &       &       &       &       &       &       &       &       &       &       &       &       &  \\
    \hline
    \end{tabular}%
  \label{cost_sens_tuning}%
\end{table*}%

Each of the considered models have a set of hyper-parameters (cost penalty $C$ and kernel parameter $\sigma$ for SVM while regularization coefficient $\lambda$ and number of hidden neurons $n_h$ for ELM) which needs tuning to suit the data set. A full grid search cross-validation is employed where the optimal combination of hyper-parameters are determined based on observed total accuracy of the classifier. The hyper-parameter tuning results are shown in Table \ref{ELM_tuning1} for ELM, Table \ref{SVM_tuning1} for SVM for the no-sampling and re-sampling cases. It can be observed that the total accuracy is generally high for SVM models compared to the ELM models. It can also be observed that by under-sampling or over-sampling the data, better accuracies can be achieved compared to the no-sampling case. Also both sampling methods give similar accuracy levels for both ELM and SVM models. However, an advantage of under-sampling can be realized in reduced computation as training is performed with a smaller subset of the training data.

\begin{figure}[]
      \centering
      \includegraphics[scale=0.45]{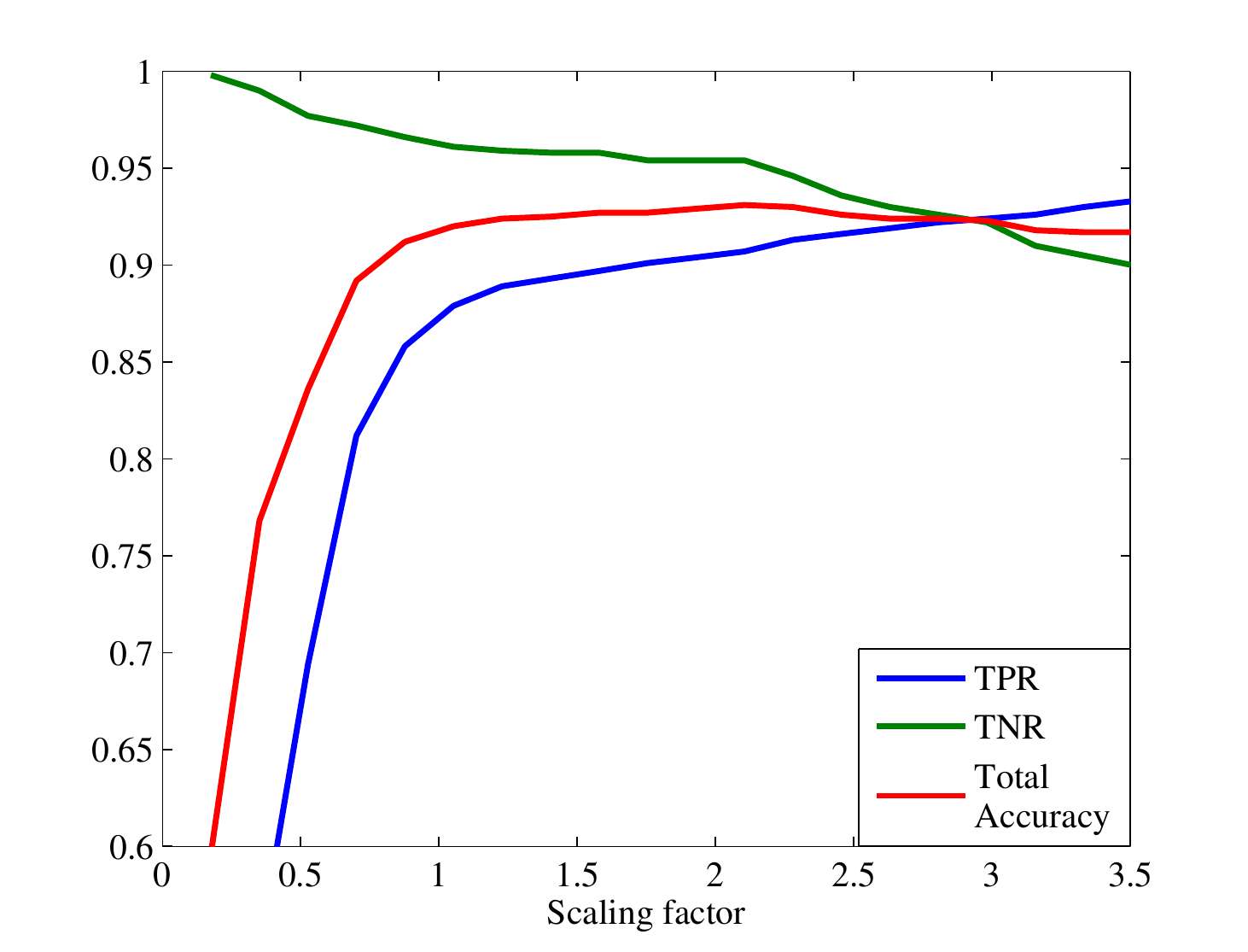}
      \caption{Sensitivity plot for TPR, TNR and Total Accuracy with scaling factor $f$ for cost-sensitive SVM.}
      \label{svm_sens}
\end{figure}

\begin{figure}[]
      \centering
      \includegraphics[scale=0.45]{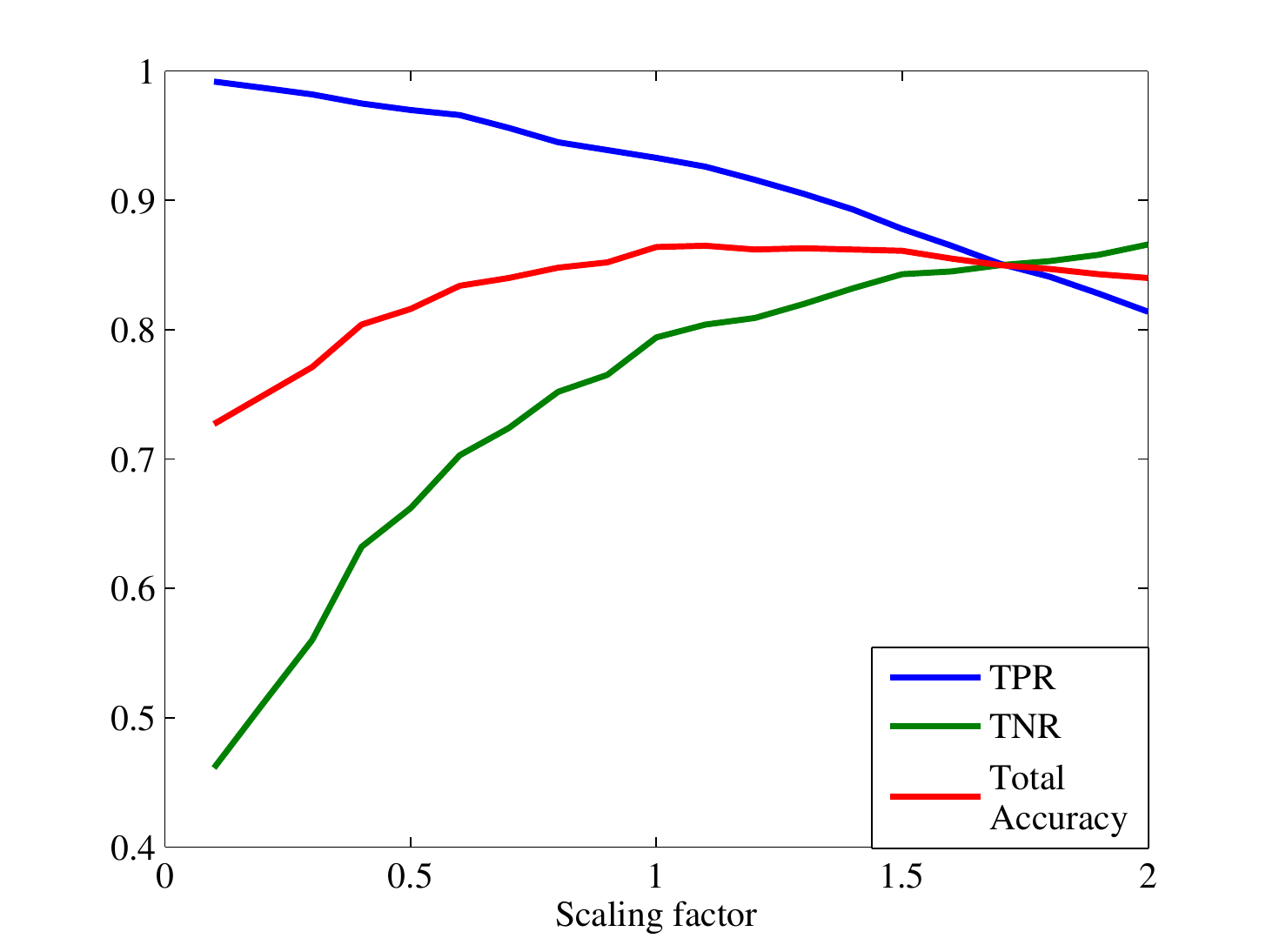}
      \caption{Sensitivity plot for TPR, TNR and Total Accuracy with scaling factor $f$ for cost-sensitive ELM.}
      \label{elm_sens}
\end{figure}

The model tuning for cost-sensitive SVM and ELM is summarized in Table \ref{cost_sens_tuning}. It can be observed that the cost-sensitive models give the desired result of accurate positive and negative class predictions without re-sampling the data. Also, the total accuracy levels are slightly higher compared to the re-sampling methods. However, it can be observed that the TPR and TNR are not close to each other for both ELM and SVM models and the same can be observed for under-sampling and over-sampling cases too indicating that it could be a limitation in the data set to classify both classes to similar accuracy. By varying the scaling factor $f$, the boundary can be perturbed to suit the application which require either high TPR or high TNR. Sensitivity plots have been shown in Figure \ref{svm_sens} and Figure \ref{elm_sens} for SVM and ELM models respectively to observe the variation of total accuracy with the scaling factor. By understanding the sensitivity of the weight factors, an optimal weight for the minority class data (combination of $r$ and $f$) can be determined.


As mentioned earlier, the SVM models have a better total accuracy compared to the ELM models. One reason could be that SVM minimizes a hinge loss while ELM minimizes a squared loss as shown in Figure \ref{app:loss_fig} and hence better regularization performance for SVM. Another reason could be that the ELM models are too simple to identify the decision boundary accurately. A possibility is to add more number of hidden neurons to the ELM models. Also, the ELM solution greatly depends on the random initialization of the input layer parameters ($W_r$ and $b_r$). In an attempt to evaluate more number of hidden neurons and different random initializations of the input layer parameters, a further experiment was conducted and results summarized in Table \ref{ELM_tab}. It can be observed that as more number of hidden neurons are added, the total accuracy improves. Also, different randomization helps in finding a compact model at a given accuracy level (compare case 2 with case 9 where 400 additional neurons are required for a negligible improvement). Hence determining an efficient way of initialization of input layer parameters is required for ELM models and will be considered in the future.

\begin{table}[]
  \centering
  \footnotesize
  \caption{Cost-sensitive ELM models with different random initialization of input layer parameters}
    \begin{tabular}{ccccccc}
    \hline
     \# & TPR   & TNR   & Total  & $\lambda$   & $n_h$    & $f$ \\
     &       &       &    Accuracy     &     &       &  \\
    \hline
    1     & 0.932 & 0.871 & 0.901 & 10.000 & 800.000 & 1.2 \\
    \textbf{2} & \textbf{0.921} & \textbf{0.879} & \textbf{0.900} & \textbf{10.000} & \textbf{600.000} & \textbf{1.3} \\
    3     & 0.933 & 0.871 & 0.902 & 10.000 & 900.000 & 1.3 \\
    4     & 0.930 & 0.863 & 0.896 & 10.000 & 600.000 & 1.2 \\
    5     & 0.934 & 0.859 & 0.896 & 10.000 & 800.000 & 1.3 \\
    6     & 0.927 & 0.871 & 0.899 & 10.000 & 800.000 & 1.2 \\
    7     & 0.939 & 0.864 & 0.902 & 10.000 & 800.000 & 1.2 \\
    8     & 0.927 & 0.869 & 0.898 & 10.000 & 700.000 & 1.1 \\
    9     & 0.929 & 0.873 & 0.901 & 10.000 & 1000.000 & 1.3 \\
    10    & 0.928 & 0.866 & 0.897 & 10.000 & 900.000 & 1.3 \\
    \hline
    \end{tabular}%
  \label{ELM_tab}%
\end{table}%

\subsection{Prediction Results} \label{results}
\begin{table*}[]
  \centering
  \footnotesize
  \caption{Summary of results for SVM and ELM models for all cases (Regular model, under-sampling, over-sampling and cost-sensitive). The results of the linear models (logistic regression and linear least squares) are also compared. The value of hyper-parameters and number of model parameters $n_p$ are also included for every model}
    \begin{tabular}{ccccccccccc}
    \hline
    &       &       &       &       &       &       &       &       &       &  \\
          & \multicolumn{4}{c}{SVM}       &    &   & \multicolumn{3}{c}{ELM}       & \multicolumn{1}{c}{} \\
          &       &       &       &       &       &       &       &       &       &  \\
          \hline
                    &       &       &       &       &       &       &       &       &       &  \\
          & Regular & Under- & Over- & Cost- &       & Regular & Under- & Over- & Cost- & Best  \\
          &  & sampling & sampling & sensitive &       &  & sampling & sampling & sensitive & ELM model \\
                    &       &       &       &       &       &       &       &       &       &  \\
    TPR   & 0.987 & 0.909 & 0.899 & 0.907 &       & 0.995 & 0.918 & 0.944 & 0.933 & 0.921 \\
    TNR   & 0.650 & 0.925 & 0.958 & 0.954 &       & 0.433 & 0.794 & 0.775 & 0.794 & 0.879 \\
    Total Accuracy & 0.819 & 0.917 & 0.928 & 0.931 &       & 0.714 & 0.856 & 0.859 & 0.864 & 0.900 \\
          &       &       &       &       &       &       &       &       &       &  \\
    $\lambda$   & -   & -   & -   & -   &       & 0.010 & 0.100 & 10.000 & 1.000 & 10.000 \\
    $n_h$    & -   & -   & -   & -   &       & 90.000 & 90.000 & 90.000 & 90.000 & 600.000 \\
    $f$  & -   & -   & -   & 2.104 &       & -   & -   & -   & 0.909 & 0.769 \\
    $C$    & 100   & 10    & 0.1   & 1     &       & -   & -   & -   & -   & - \\
    $\sigma$ & 1     & 1     & 10    & 10    &       & -   & -   & -   & -   & - \\
    $n_p$    & 33696 & 16965 & 236691 & 120900 &       & 3690  & 3690  & 3690  & 3690  & 24600 \\
          &       &       &       &       &       &       &       &       &       &  \\
    \hline
          &       &       &       &       &       &       &       &       &       &  \\
                &       &       &       &       &       &       &       &       &       &  \\
                            &       &       &       &       &       &       &       &       &       &  \\
          \hline
          &       &       &       &       &       &       &       &       &       &  \\
          & \multicolumn{3}{c}{Logistic Regression} &       &       & \multicolumn{4}{c}{Linear LS} &  \\
          &       &       &       &       &       &       &       &       &       &  \\
          \hline
                &       &       &       &       &       &       &       &       &       &  \\
          & Regular & Under- & Over- &       &       & Regular & Under- & Over- & Cost-&  \\
          &  & sampling & sampling &       &       &  & sampling & sampling & sensitive &  \\
                    &       &       &       &       &       &       &       &       &       &  \\
    TPR   & 0.995 & 0.911 & 0.928 &       &       & 0.996 & 0.941 & 0.955 & 0.875 &  \\
    TNR   & 0.441 & 0.791 & 0.786 &       &       & 0.389 & 0.704 & 0.699 & 0.828 &  \\
    Total Accuracy & 0.718 & 0.851 & 0.857 &       &       & 0.692 & 0.822 & 0.827 & 0.852 &  \\
    $n_p$    & 40 & 40 & 40 &  &       & 40  & 40  & 40  & 40  &  \\
          &       &       &       &       &       &       &       &       &       &  \\
          \hline
    \end{tabular}%
  \label{summary_table}%
\end{table*}%

The models developed using SVM and ELM are compared against baseline linear models and the results summarized in Table \ref{summary_table}. The case 2 model in Table \ref{ELM_tab} is considered as the best ELM model and included in the summary table. From the modeling results, it can be observed that both re-sampling methods (under-sampling and over-sampling) as well as cost-sensitive classification are suitable for the problem considered in this work. The nonlinear models result in better accuracies compared to the linear models indicating that the HCCI boundary is a nonlinear system and that nonlinear classification methods are necessary. However the cost paid for selecting a nonlinear model is the additional computation and memory required and the tradeoff can be evaluated specific to the application based on the importance of having accurate versus having low complexity models. For instance, the number of parameters required to identify the classification boundary for different models is summarized in Table \ref{summary_table}. It is obvious that the decision boundary identified by linear models (under-parameterized) is very simple and does not capture the right behavior. SVM and ELM models on the other hand requires a large number of parameters in an attempt to capture more complex behavior. SVM is a non-parametric model and hence the number of parameters grows with number of training data. ELM on the other hand is a parametric model and hence the number of parameters are fixed and hence requires about 80\% less number of parameters for an inaccuracy of 3\%. This is a major drawback of SVM for applications onboard the engine ECU which is limited in memory and computation.

For the engine problem considered here, it was observed that both SVM and ELM are capable of identifying the stable and unstable boundary of HCCI from experimental data. However, an important criteria for HCCI engine application is that the models must be able to be adapted online. As HCCI combustion is sensitive to other variables like engine speed, ambient temperature, pressure and humidity levels etc. and hence experiments and data size might increase rapidly. Also, performing an experiment that  covers the entire region of operation is infeasible. Hence a system that can learn and adapt in a sequential manner is of extreme importance. Hence even though SVM outperformed ELM in terms of accuracy, ELM requires relatively less number of parameters and is simple and efficient in implementation for on-line learning and chosen as suitable for the application in hand.

\begin{figure}[]
      \centering
      \subfloat[]{\label{elm_pred_zoom}\includegraphics[scale=0.6]{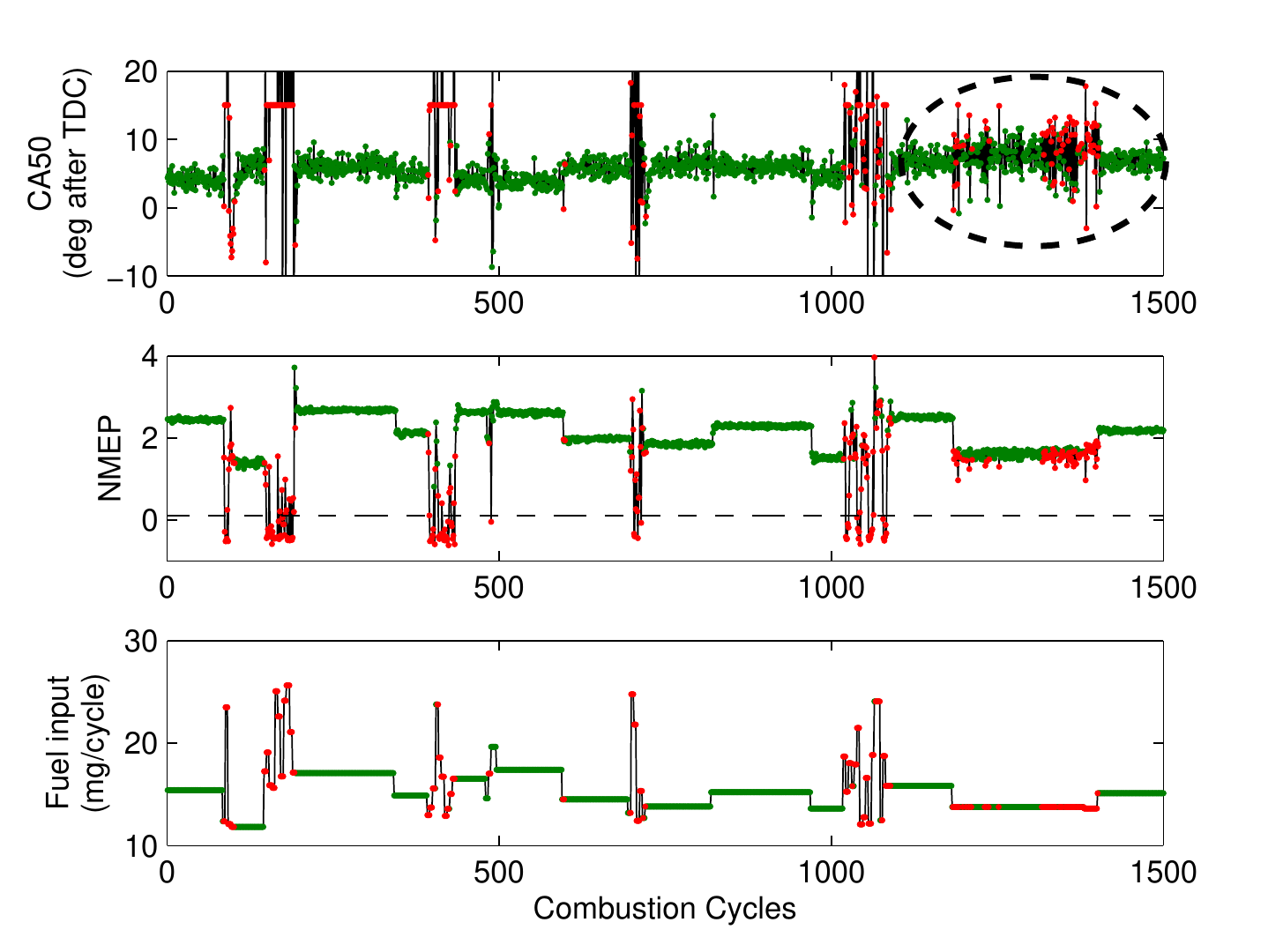}}\\
      \subfloat[]{\label{svm_pred_zoom}\includegraphics[scale=0.6]{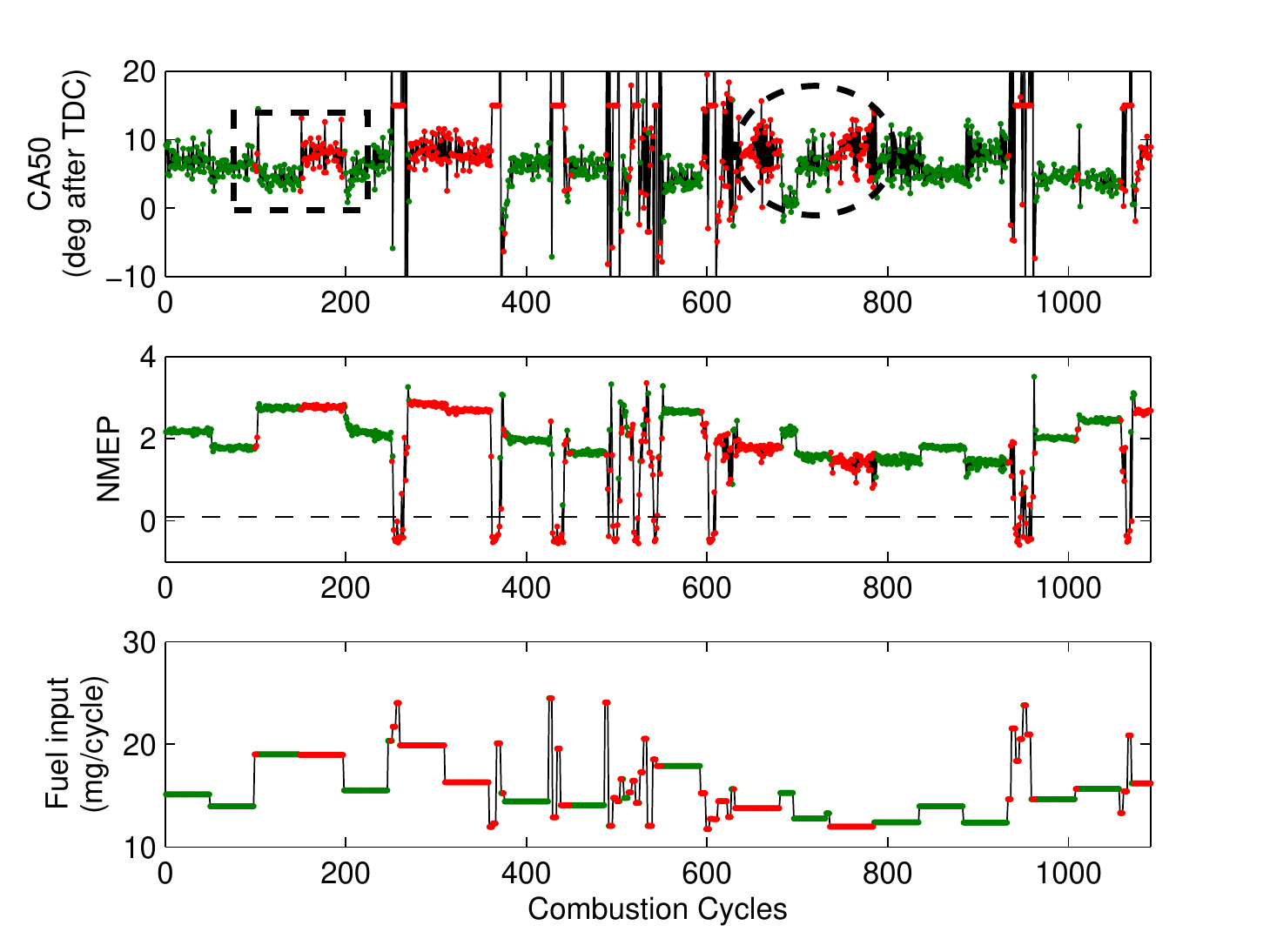}}\\
                  \subfloat[]{\label{svm_pred_zoom}\includegraphics[scale=0.6]{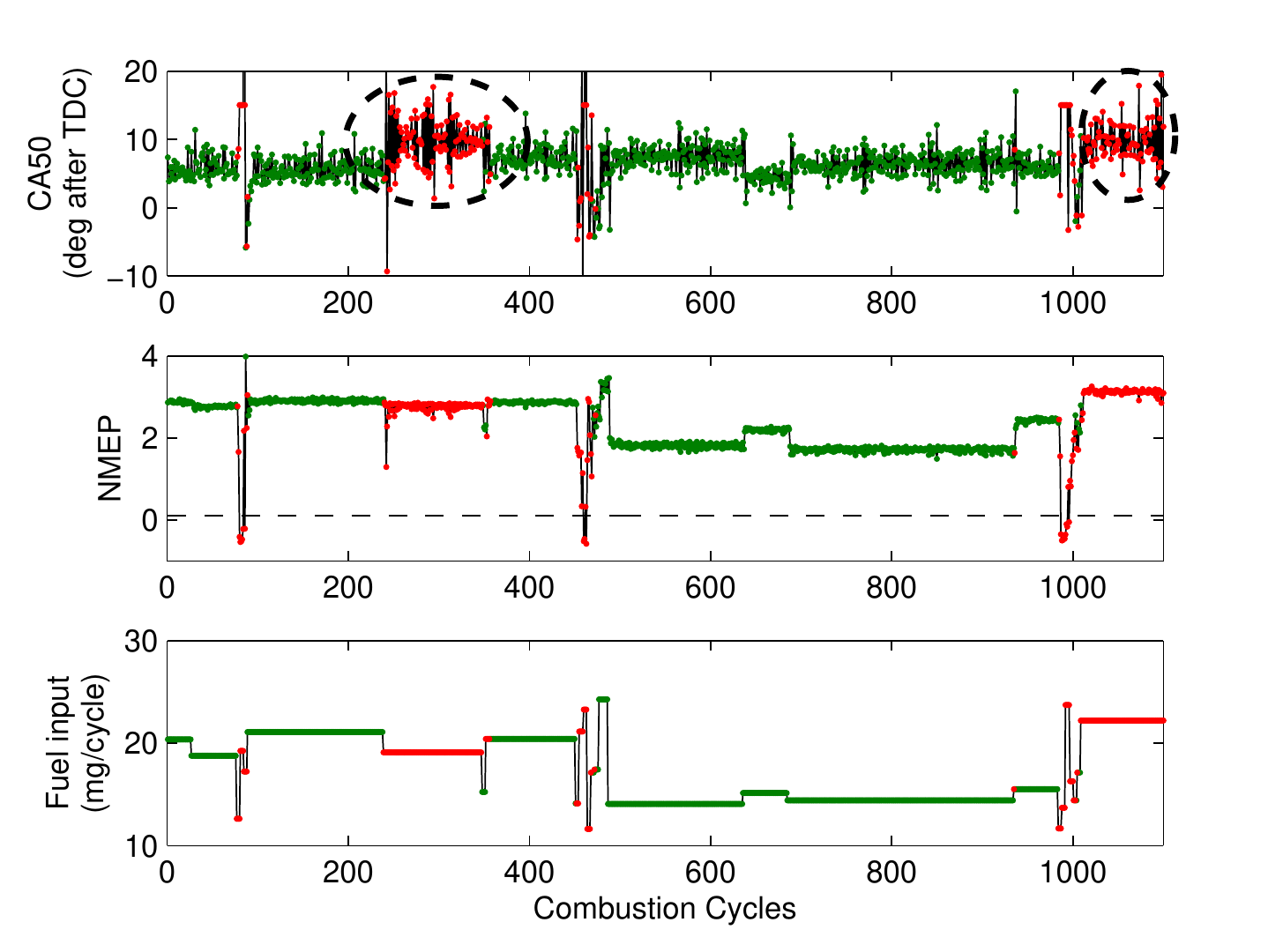}}
      \caption{Prediction results of the cost-sensitive ELM showing CA50, IMEP and one input variable (fueling) for 3 unseen data sets. The color code indicates model prediction - green (and red) indicate stable (and unstable) prediction by the model. The dotted line in the IMEP plot indicates misfire limit, dotted ellipse in CA50 plot indicates high variability instability mode while dotted rectangle indicates a wrong predictions by model.}
      \label{elm_pred}
\end{figure}

\begin{figure}[]
      \centering
      \subfloat[]{\label{elm_pred_zoom}\includegraphics[scale=0.6]{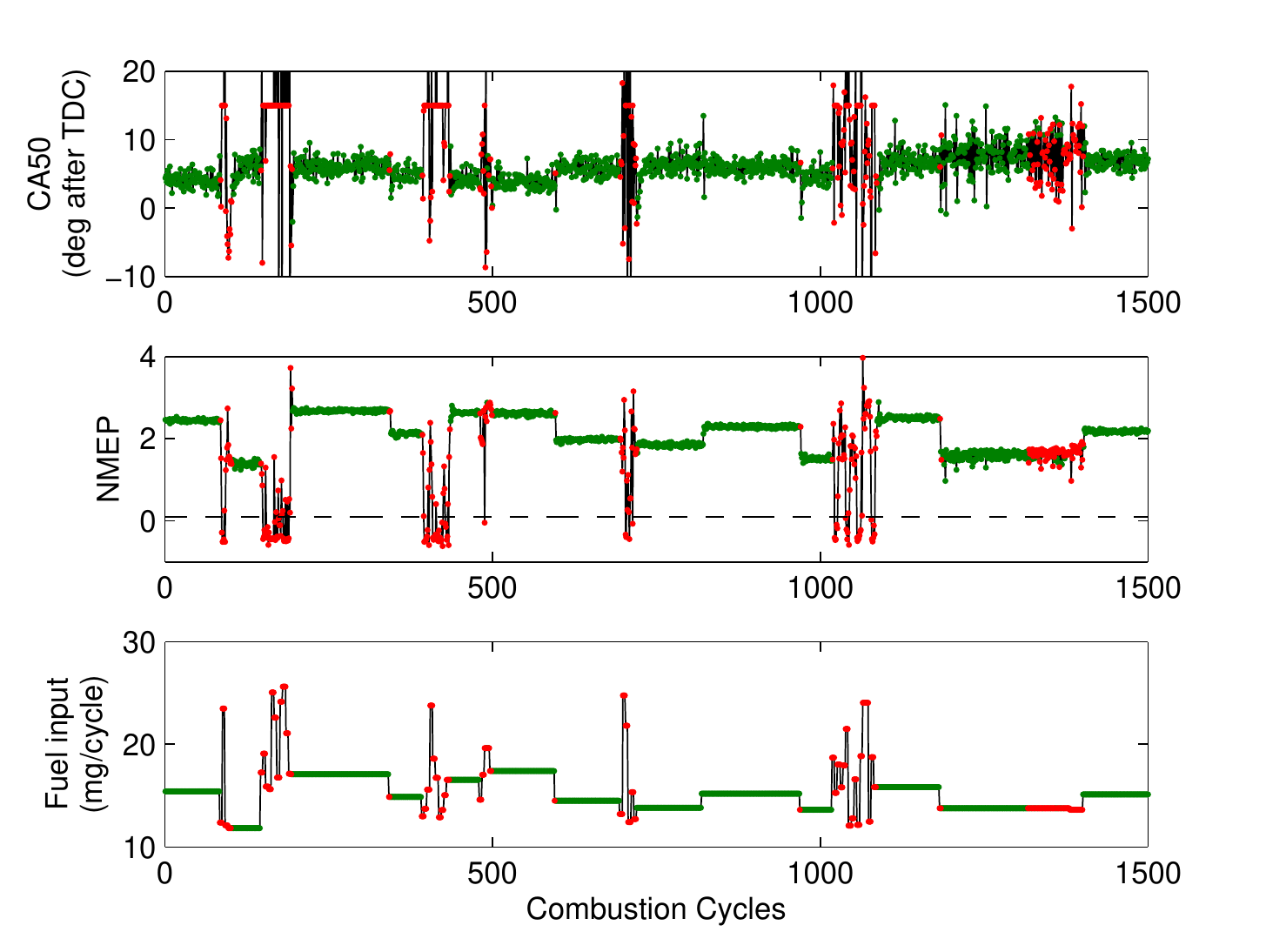}}\\
      \subfloat[]{\label{svm_pred_zoom}\includegraphics[scale=0.6]{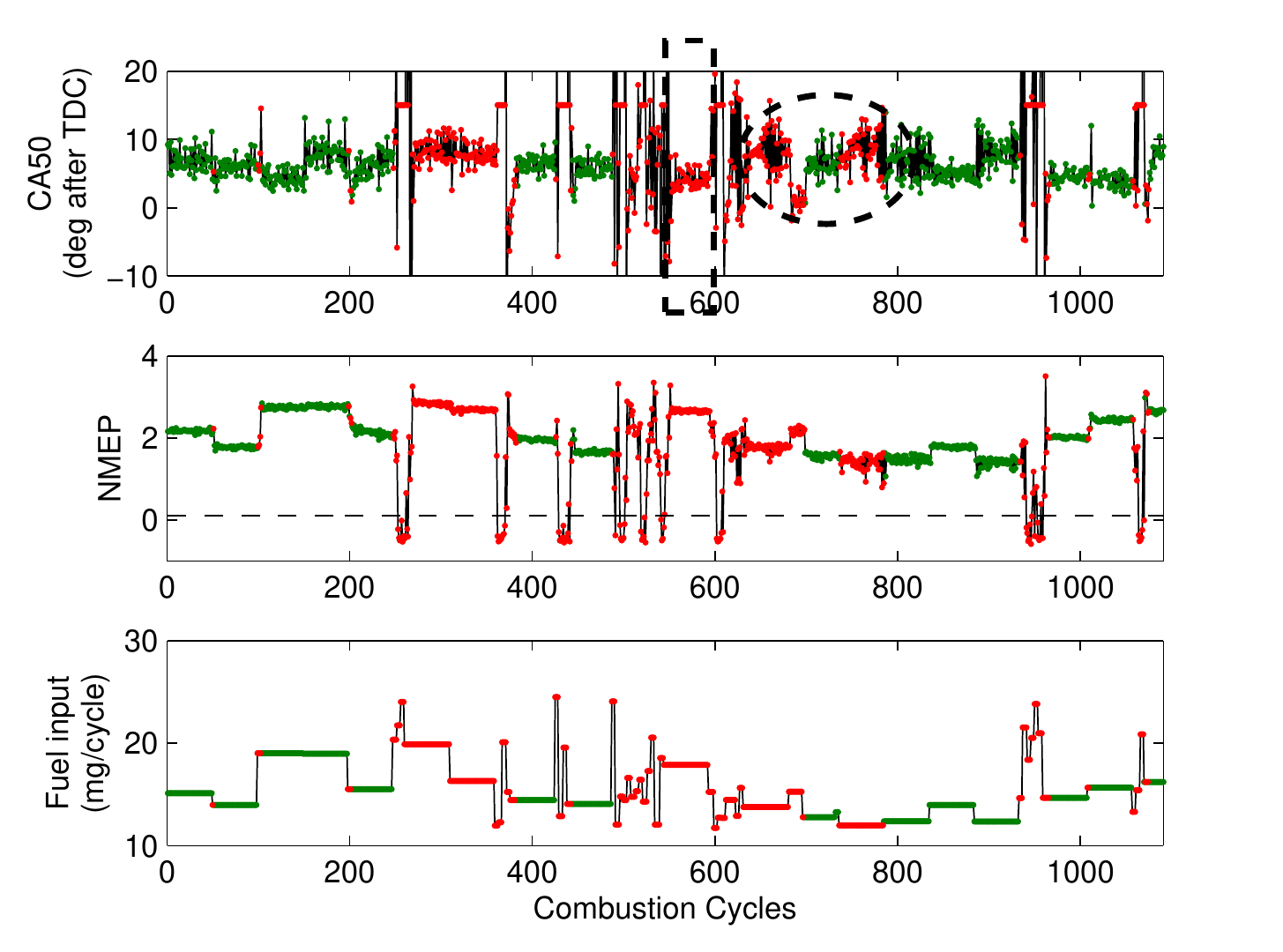}}\\
                  \subfloat[]{\label{svm_pred_zoom}\includegraphics[scale=0.6]{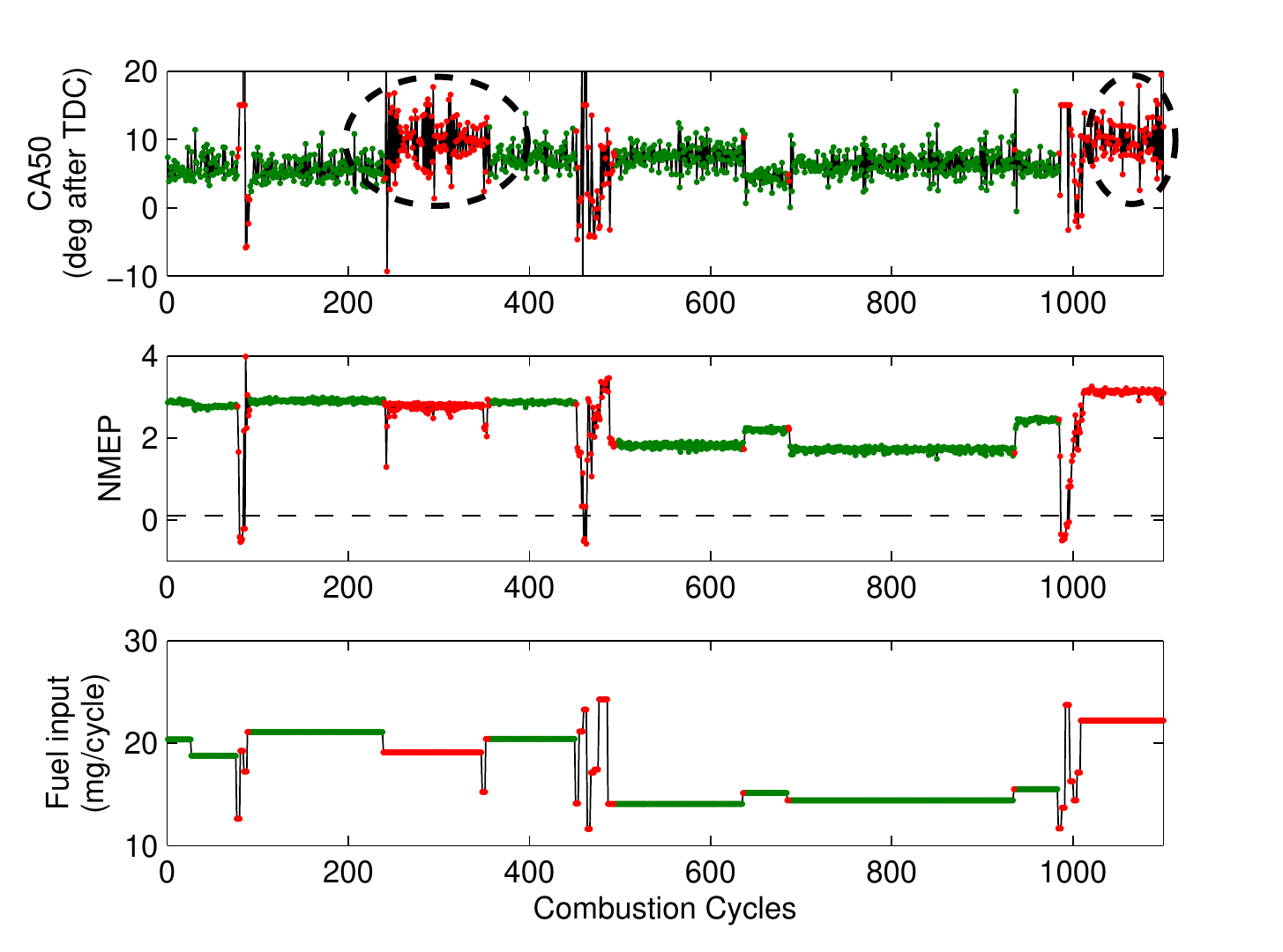}}
      \caption{Prediction results of the cost-sensitive SVM showing CA50, IMEP and one input variable (fueling) for 3 unseen data sets. The color code indicates model prediction - green (and red) indicate stable (and unstable) prediction by the model. The dotted line in the IMEP plot indicates misfire limit, dotted ellipse in CA50 plot indicates high variability instability mode while dotted rectangle indicates a wrong predictions by model.}
      \label{svm_pred}
\end{figure}

The most accurate model using ELM and SVM are used to make predictions on unseen engine inputs and predictions are summarized in Figure \ref{elm_pred} and Figure \ref{svm_pred} respectively while quantitative results are included in Table \ref{summary_table}. Color codes are used to represent the predictions - a red marker on the plots indicate the model's prediction being "unstable" while a green marker indicates prediction being "stable". It can be seen from the plots that both models predict the HCCI instabilities well. Dotted lines are shown as references for stable operation (Data falling outside the dotted lines in CA50 plot or data falling below the dotted line in NMEP plot indicate instability).

In order to get a closer look, a subsection of the above figures are plotted in Figure \ref{elm_pred_zoom} and Figure \ref{svm_pred_zoom} for ELM and SVM respectively. It can be observed from Figure \ref{elm_pred_zoom} that the input sample (at cycle 34) along with other input measurements is predicted unstable and rightly so, the following cycles are unstable as indicated by CA50 overshooting 10 degree after TDC. This is the primary goal of the proposed method that predicts if the engine operation is stable/unstable at time $k+1$, given the measurements up to time $k$. Hence using the model, any given input actuator setting and history of measurements of key combustion variables, it would be possible to predict if the subsequent set of combustion cycles misfire or not. A similar plot can be shown for SVM in Figure \ref{svm_pred_zoom} which has a slightly different prediction compared to the ELM model. For instance, SVM predicts most of the region beyond cycle 34 to be unstable but ELM predicts to be stable between cycles 75 and 95. However, these cycles do not fall into completely stable or completely unstable and hence might be in the fuzzy region between the two classes. Such predictions are to be expected from both models as training was not performed using a comprehensive data set that had a dense distribution of data in both classes.

\begin{figure}[]
      \centering
      \subfloat[Cost-sensitive ELM]{\label{elm_pred_zoom}\includegraphics[scale=0.5]{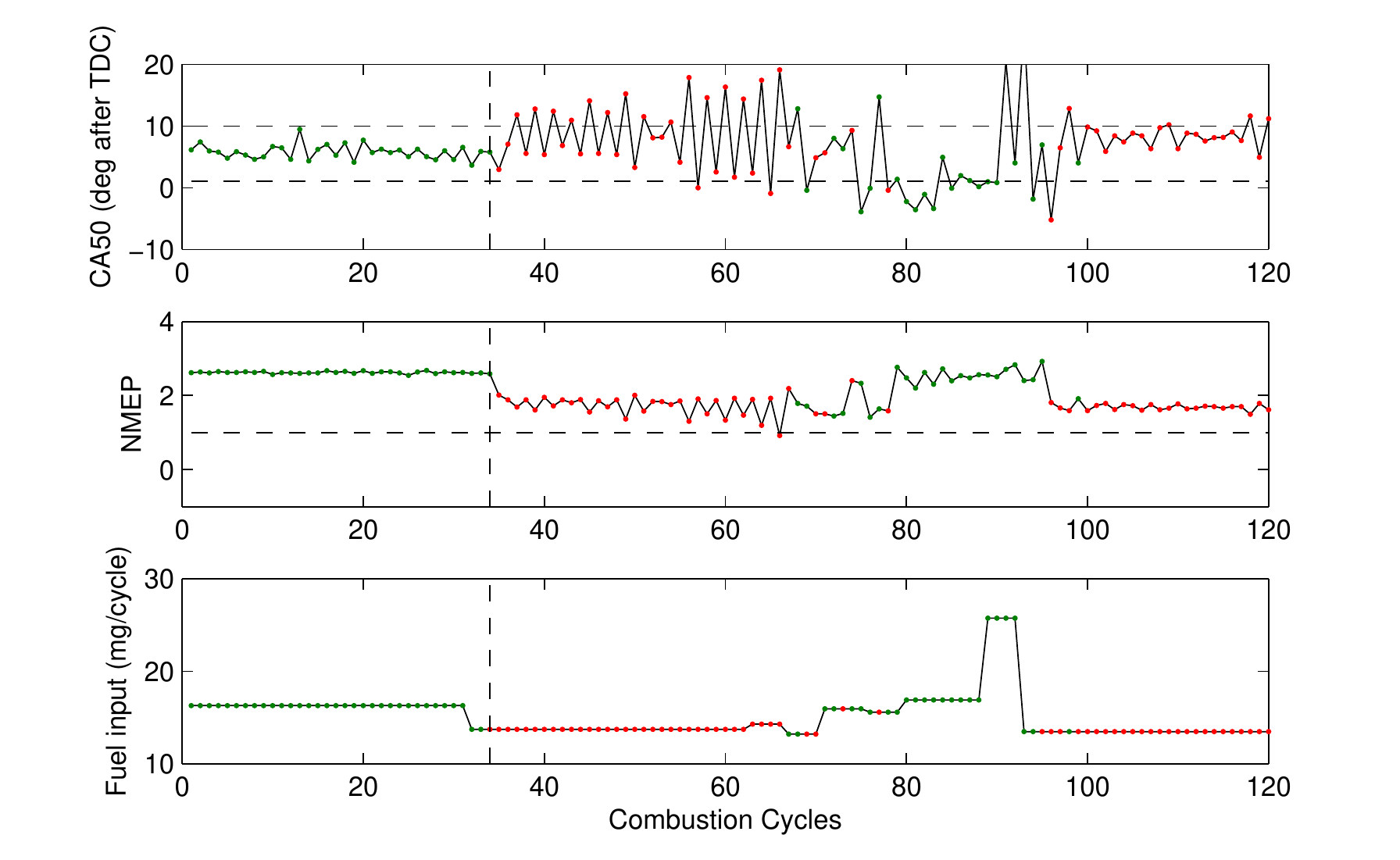}}\\
      \subfloat[Cost-sensitive SVM]{\label{svm_pred_zoom}\includegraphics[scale=0.5]{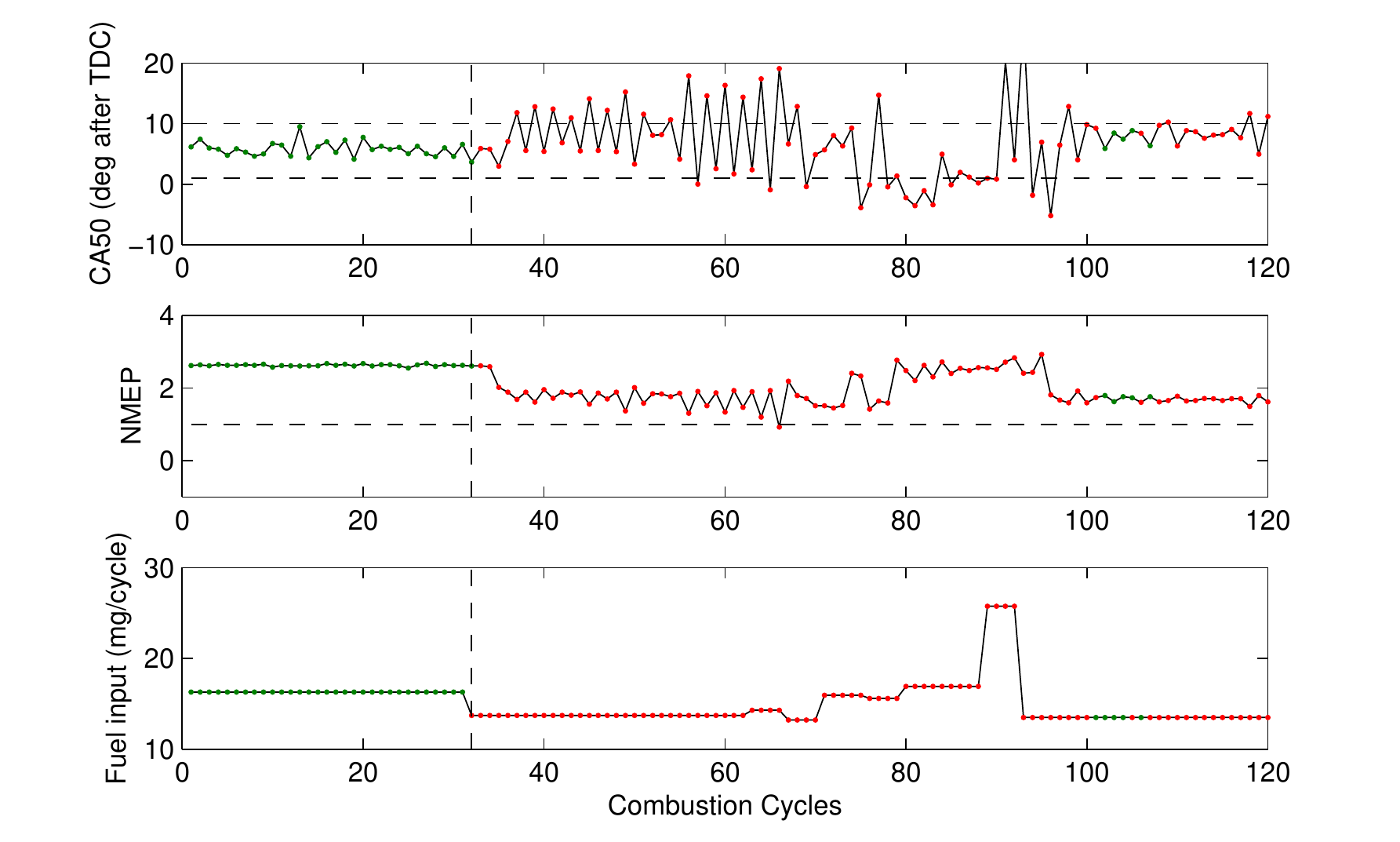}}
      \caption{A small subset of prediction results of the cost-sensitive ELM and SVM showing CA50, IMEP and one input variable to compare predictions in perspective to input variables. The green points indicate stable operation while red points indicate unstable operation.}
      \label{zoomed_figs}
\end{figure}

\section{Conclusion} \label{concl}
Complex and highly sensitive systems such as HCCI engines have a narrow region of stable operation and it is important to gain knowledge about the operating envelope for diagnostics and controls development. In this paper, a novel solution using soft computing has been developed that predicts the future combustion events based on past and present measurements along with excitation inputs. An imbalanced classification problem has been formulated and solved using linear and nonlinear methods such as logistic regression, linear regression, SVM and ELM.

A comparison of data re-sampling methods and cost-sensitive learning approaches have been performed and results summarized. Re-sampling methods are found to work well but cost-sensitive methods have a slightly better accuracy and avoid artificial modifications in the data distributions. A modification to the ELM algorithm has been made by weighting the minority class data more to handle the imbalance in the data set. The cost-sensitive SVM classifier outperforms the other algorithms in terms of accuracy but requires a large fraction of the data to be stored for predictions, typical of non-parametric methods. ELM (and its variants) results in an inferior accuracy compared to SVM (and corresponding variants) but preferred over SVM as the ELM model is much simpler (less number of parameters) and is more suitable for online learning for a memory-limited, real-time application such as the HCCI engine. Future work will involve identifying more unstable modes of HCCI and perform online model adaptation using ELM's online learning algorithm to improve accuracy and explore unexcited engine operation. Further, application of the developed model towards a closed system identification and controller development will be explored in the future.


%

\section*{Acknowledgment}
This material\footnote{\scriptsize Disclaimer: This report was prepared as an account of work sponsored by an agency of the United States Government.  Neither the United States Government nor any agency thereof, nor any of their employees, makes any warranty, express or implied, or assumes any legal liability or responsibility for the accuracy, completeness, or usefulness of any information, apparatus, product, or process disclosed, or represents that its use would not infringe privately owned rights.  Reference herein to any specific commercial product, process, or service by trade name, trademark, manufacturer, or otherwise does not necessarily constitute or imply its endorsement, recommendation, or favoring by the United States Government or any agency thereof.  The views and opinions of authors expressed herein do not necessarily state or reflect those of the United States Government or any agency thereof.} is based upon work supported by the Department of Energy and performed as a part of the ACCESS project consortium (Robert Bosch LLC, AVL Inc., Emitec Inc.) under the direction of PI Hakan Yilmaz, Robert Bosch, LLC.

\appendices
\section{Loss functions}\label{app:loss}
\begin{figure}[]
      \centering
      \includegraphics[scale=0.5]{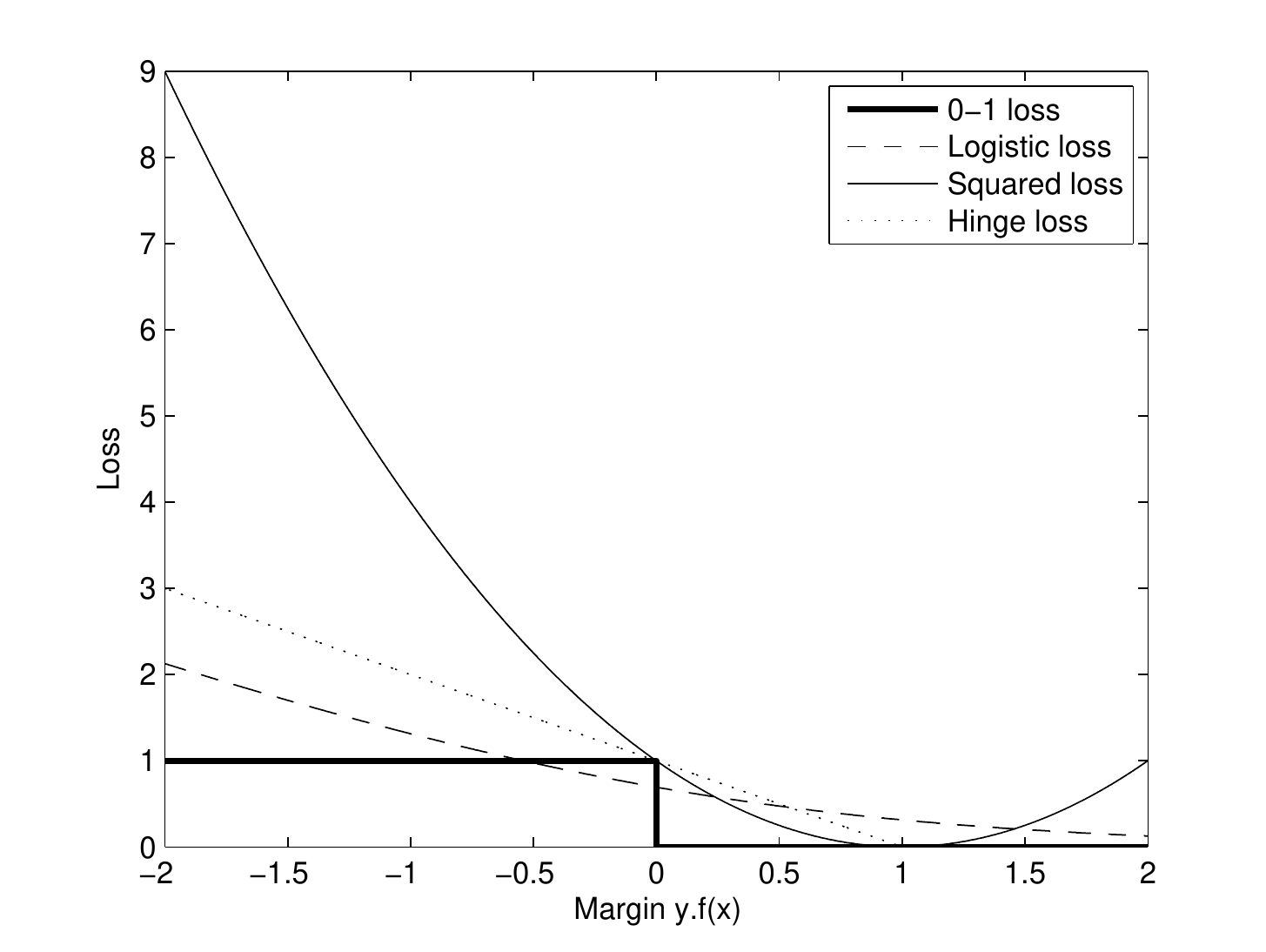}
      \caption{A comparison of the loss functions of the algorithms used in this paper with the baseline 0-1 error standard.}
      \label{app:loss_fig}
\end{figure}

A plot showing the different loss functions used for classification can be shown in Fig. \ref{app:loss_fig}. The 0-1 loss is the most efficient loss function for binary classification but is non-differentiable and hence not used in developing algorithms. A logistic loss is given by equation \eqref{logloss} is used in logistic regression algorithm while a hinge loss (equation \eqref{hingeloss}) and squared loss (equation \eqref{sqloss}) are implemented in SVM and ELM algorithms respectively.
\begin{eqnarray}
L_{logistic}&=&\log (1+e^{-y f(x)}) \label{logloss} \\
L_{hinge}&=&max(0,1-y f(x)) \label{hingeloss} \\
L_{squared}&=&(1-y f(x))^2 \label{sqloss}
\end{eqnarray}

\section{Logistic Regression} \label{app:logreg}
The goal of logistic regression is to determine $\beta$ such that $P(Y|X,\beta)$ is maximized, i.e, to solve the following optimization problem
\begin{equation}\label{}
\beta=\arg\max_\beta P(Y|X,\beta)=\arg \max_\beta \Pi_{i=1}^N P(y_i|x_i,\beta)
\end{equation}
Expressing the above in log likelihood form, the optimization problem becomes
\begin{eqnarray}\label{}
\nonumber \beta &=& \arg\max_\beta \log\{\Pi_{i=1}^N P(y_i|x_i,\beta)\} \\
\nonumber       &=& \arg\max_\beta \sum_{i=1}^N \log P(y_i|x_i,\beta) \\
\nonumber       &=& \arg\max_\beta \sum_{i=1}^N \log  \frac{1}{1+e^{-y(\beta_1^T x+\beta_0)}} \\
\nonumber       &=& \arg\max_\beta -\sum_{i=1}^N \log  1+e^{-y(\beta_1^T x+\beta_0)} \\
      &=& \arg\min_\beta \sum_{i=1}^N \log  \left(1+e^{-y(\beta_1^T x+\beta_0)}\right)
\end{eqnarray}


\ifCLASSOPTIONcaptionsoff
  \newpage
\fi

\bibliographystyle{IEEEtran}
\bibliography{IEEEabrv,misfire}

\end{document}